\pdfoutput=1
\documentclass[11pt]{article}
\usepackage{times,latexsym}
\usepackage{url}
\usepackage[T1]{fontenc}

\usepackage[acceptedWithA]{tacl2021v1}

\usepackage{times}
\usepackage{latexsym}
\usepackage{xcolor}
\usepackage{tabularx,booktabs}
\usepackage{booktabs}
\usepackage{multirow}
\usepackage{hyperref}
\usepackage{cleveref}
\usepackage{xspace}
\usepackage[T1]{fontenc}
\usepackage[utf8]{inputenc}

\usepackage{microtype}
\usepackage{subcaption}
\usepackage{graphicx}
\usepackage{inconsolata}

\usepackage{color, colortbl}

\newcommand{\rparagraph}[1]{\vspace{1mm}\noindent\textbf{#1.}}
\newcommand{\rrparagraph}[1]{\vspace{0.4mm}\noindent\textit{#1.}}

\newcommand{\rparagraphnodot}[1]{\vspace{1mm}\noindent\textbf{#1}}
\newcommand{\sparagraphnodot}[1]{\vspace{0.0mm}\noindent\textbf{#1}}

\definecolor{Gray}{gray}{0.92}

\usepackage{todonotes}
\makeatletter
\newcommand*\iftodonotes{\if@todonotes@disabled\expandafter\@secondoftwo\else\expandafter\@firstoftwo\fi}
\makeatother

\definecolor{genielime}{rgb}{0.9,1,0.3}

\definecolor{ivanc}{rgb}{0.0,0.8,0.9}

\title{Analyzing and Adapting Large Language Models for Few-Shot Multilingual NLU: Are We There Yet?}

\author{Evgeniia Razumovskaia~~~~~~Ivan Vuli\'{c}~~~~~~ Anna Korhonen \\
        Language Technology Lab, University of Cambridge, UK\\ \texttt{\{er563, iv250, alk23\}@cam.ac.uk}}
\begin{document}
\maketitle
\begin{abstract}
Supervised fine-tuning (SFT), supervised instruction tuning (SIT) and in-context learning (ICL) are three alternative, \textit{de facto} standard approaches to few-shot learning. ICL has gained popularity recently with the advent of LLMs due to its simplicity and sample efficiency. Prior research has conducted only limited investigation into how these approaches work for \textit{multilingual} few-shot learning, and the focus so far has been mostly on their performance. In this work, we present an extensive and systematic comparison of the three approaches, testing them on 6 high- and low-resource languages, three different NLU tasks, and a myriad of language and domain setups. Importantly, performance is only one aspect of the comparison, where we also analyse the approaches through the optics of their computational, inference and financial costs. Our observations show that supervised instruction tuning has the best trade-off between performance and resource requirements. As another contribution, we analyse the impact of target language adaptation of pretrained LLMs and find that the standard adaptation approaches can (superficially) improve target language generation capabilities, but language understanding elicited through ICL does not improve and remains limited, with low scores especially for low-resource languages. 
\end{abstract}

\section{Introduction and Motivation}\label{sec:intro}
Recent advances in data-efficient, few-shot learning have been crucial for increasing and promoting language inclusiveness of NLP technology \cite{devlin-etal-2019-bert, conneau-etal-2020-unsupervised, imanigooghari-etal-2023-glot500}, substantially lowering the dataset size-related `entry point' for a new language. This was made possible by pretrained language models which can generalise to a new task or language from the knowledge stored in their parameters complemented with only a handful of in-task data. 

%\captionsetup[subfigure]{oneside,margin={1cm}}
\captionsetup[subfigure]{oneside,margin={-4.5cm,1cm},skip=-1pt}
\begin{figure}[!t]
\centering
\begin{subfigure}[b]{0.34\textwidth}
   \includegraphics[width=1\linewidth]{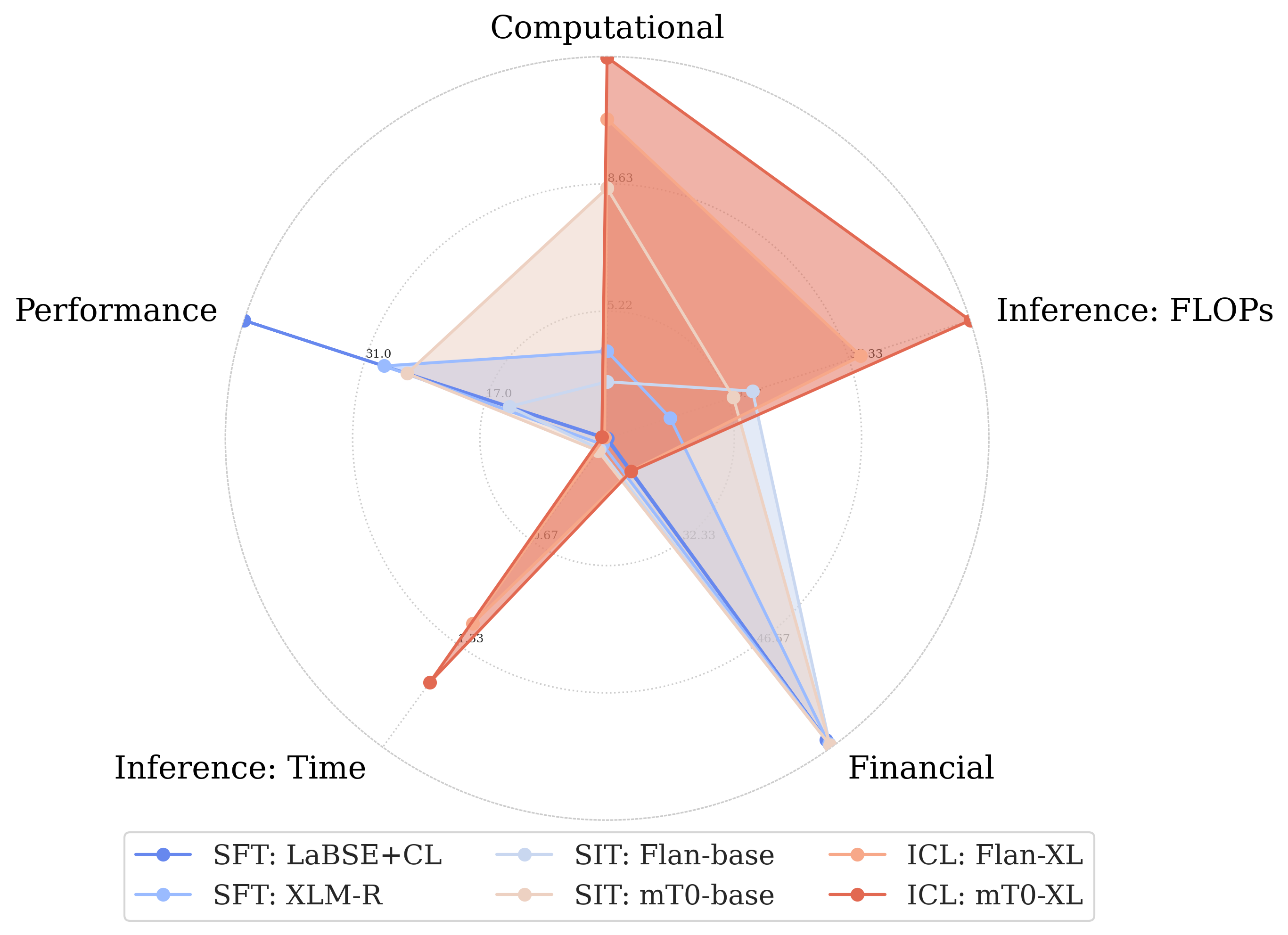}
   \caption{Amharic}
   \label{fig:comparison_practicalities_amh} 
\end{subfigure} \quad
\begin{subfigure}[b]{0.34\textwidth}
   \includegraphics[width=1\linewidth]{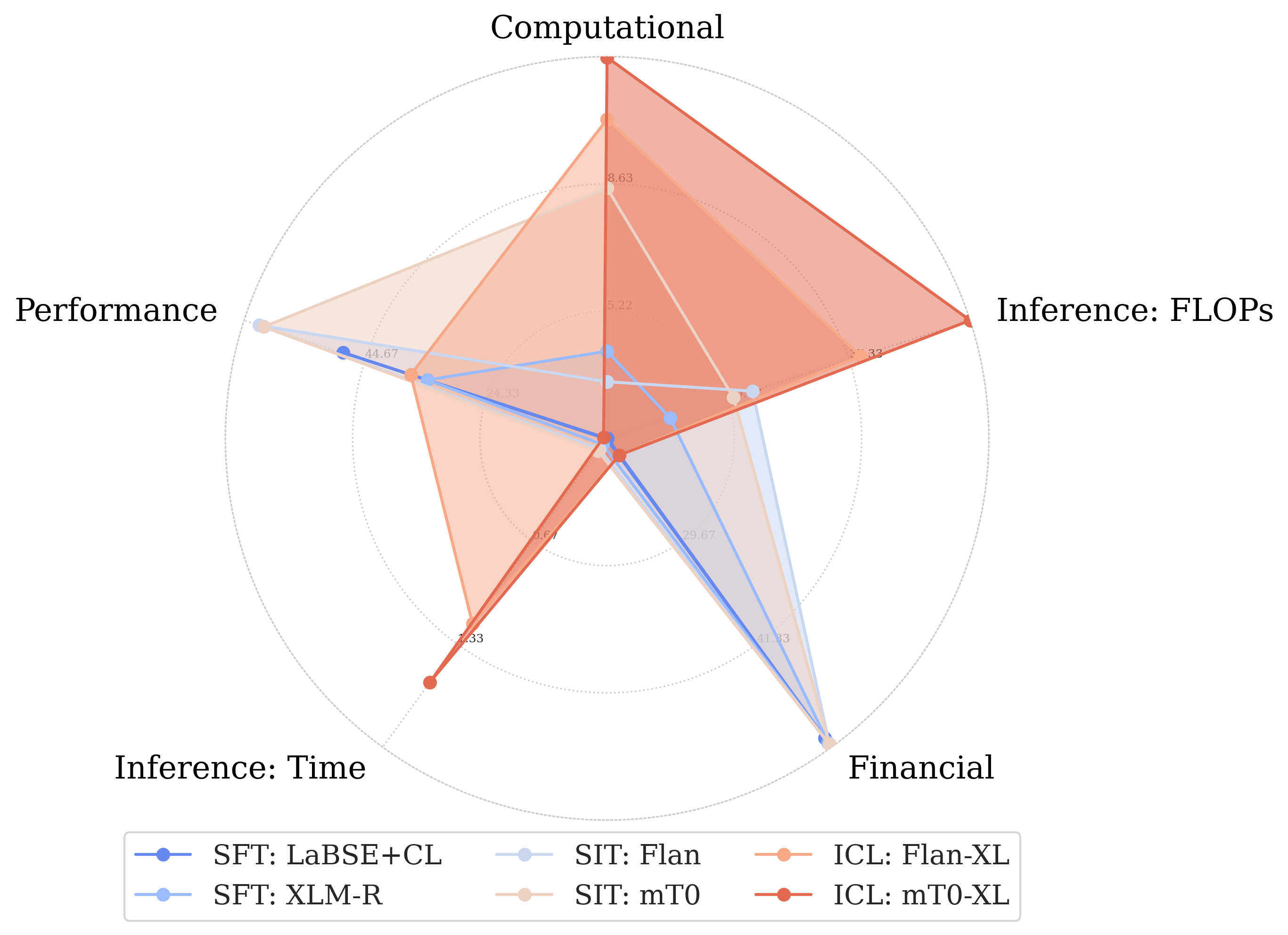}
   \caption{Spanish}
   \label{fig:comparison_practicalities_es}
\end{subfigure}
\vspace{-1mm}
\caption{Comparison of practical aspects of different learning paradigms (\S\ref{subsec:learning_paradigms}) in the intent detection task from Multi3NLU++~\cite{moghe-etal-2023-multi3nlu}, with exactly the same data setup, for Amharic and Spanish. In-context learning (ICL) has low performance and high inference and computational costs while being comparatively inexpensive. Supervised fine-tuning (SFT) and supervised instruction-tuning (SIT), on the other hand, have a larger financial cost but they are much more efficient in terms of inference aspects and computational resources while also performing much better both for Amharic as a representative low-resource language (\ref{fig:comparison_practicalities_amh}) and Spanish as a high-resource language (\ref{fig:comparison_practicalities_es}).}\label{fig:comparison_practicalities}
\vspace{-1.5mm}
\end{figure}
%\ivan{Figure 1 is still not fully fine: why Flan and mT0 without the model size with SIT? Also increase the font of those legends.}

The standard approaches for such few-shot adaptations are \textit{Supervised Fine-Tuning (SFT)}, which also subsumes more recent \textit{Supervised Instruction-Tuning (SIT)}, and \textit{In-Context Learning (ICL)}. SFT and SIT use knowledge in pretrained model parameters for initialisation and then adapt the parameters to a \textit{language-task} combination via \textit{supervised training} on available, even if scarce, resources. Importantly, they yield a model specialised for a single \textit{language-task} combination and can get increasingly better at the task if a larger training dataset becomes available. ICL, in contrast, uses one model `as is' to complete any task, without any parameter adaptation or fine-tuning. Instead, the model is adapted via prompting: given an explanation of a task (i.e., \textit{instruction}) and a set of `training' examples (i.e., annotated \textit{demonstrations}), the model is tasked to generate the label for every input \citep{radford2019-in-context-learning}. Due to the model's context size, the number of demonstrations used in the input is limited, meaning that the ICL performance is capped by the model's pretraining and the demonstrations that fit into the input context. 

Existing generative models used for ICL (termed \textit{Large Language Models, or LLMs} henceforth) are usually pretrained in an English-centric manner with the vast majority of the pretraining corpus in English and only limited coverage of other languages \cite{sitaram2023everything}, even with `accidentally encountered' bilingual and translation data \citep{briakou-etal-2023-searching}. As a result, current LLMs are very far from serving the world's languages equally: while demonstrating impressive ICL results in English \citep{mishra2022cross}, they still face difficulties when transferring to other languages \citep{winata2021language,tanwar-etal-2023-multilingual}, especially low-resource ones \cite{ojo2023good}. In contrast, a number of encoder and encoder-decoder models, such as XLM-R \citep{conneau-etal-2020-unsupervised} or mT5 \citep{xue-etal-2021-mt5}, used for initialisation in SFT are pretrained with much wider language coverage\footnote{For instance, XLM-R \cite{conneau-etal-2020-unsupervised}, mBERT \cite{devlin-etal-2019-bert}, LaBSE \cite{feng-etal-2022-labse}, and mT5 \cite{xue-etal-2021-mt5} cover $\sim$100 languages at pretraining (albeit with different pretraining data amounts), while Glot500 \cite{imanigooghari-etal-2023-glot500} covers up to 500 languages.} (termed \textit{multilingually Pretrained Language Models, or mPLMs}) \cite{conneau-etal-2020-unsupervised, imanigooghari-etal-2023-glot500}. This property enables sample-efficient transfer and adaptation of natural language understanding (NLU) models to a much larger array of languages~\cite{ansell-etal-2021-mad-g} than what is supported by ICL-based LLMs.
%\ivan{This claim is too strong, smooth it out?} 

%\ivan{We should also cite some work that tried to adapt them multilingually somewhere, e.g., https://arxiv.org/abs/2204.07580, https://arxiv.org/abs/2401.01854}especially the low-resourced ones \citep{ojo2023good}.

While SFT, SIT and ICL are comparable approaches for few-shot multilingual NLU, there has been little attention drawn to which of the techniques works better \textit{in practice}. Therefore, this paper aims to delve deeper into analysing a variety of factors which critically impact effective use of either from a more practical point of view. Our first aim is to provide answers to the following question:

\vspace{0.5mm}
\noindent \textit{(Q1) Given the same annotated examples, which of the approaches is better \textbf{in practice}?}
\vspace{0.5mm}

\noindent In particular, the sometimes vague term `practice' in our work comprises the following crucial aspects: \textbf{1)} sample efficiency (i.e., \textit{`data cost'}); \textbf{2)} computational requirements (\textit{`computational cost'}); \textbf{3)} latency (\textit{`inference cost'}); and \textbf{4)} overall financial or `economic' cost. Prior work has been mainly focused on benchmarking ICL on subgroups of languages \cite{ojo2023good} and only considered and optimised task performance of the models as the ultimate comparison criterion. In contrast, our work presents an extensive analysis evaluating cross-lingual capabilities of SFT and ICL both on high and low-resource languages, considering not only the task performance but also the above listed practical aspects, as illustrated in Figure~\ref{fig:comparison_practicalities}.

%such as: a) sample efficiency; b) computational requirements; c) latency; and d) cost of each of the approaches. 

Furthermore, prior work has demonstrated the effectiveness of parameter-efficient fine-tuning (PEFT) to improve the model's cross-task capabilities and to promote aspects of its generation abilities (e.g., open-domain chat; \citeauthor{dettmers2023qlora}, \citeyear{dettmers2023qlora}). In this work, we also analyse how language adaptation of LLMs `beyond English' impacts their NLU and NLG performance in a target language. %\ivan{Will we also provide some results testing their NLG capabilities?} \genie{yes :) } 
This gives rise to another core research question:

\vspace{0.4mm}
\noindent \textit{(Q2) Given the benefits of ICL as a learning paradigm (but its inferior performance in comparison to SFT), could we use the standard \textbf{adaptation strategies} to improve LLMs' generation and understanding capabilities in other languages?}
\vspace{0.4mm}

To our knowledge, this is the first work analysing how multilingual NLU capabilities of ICL with LLMs are effected by their target language adaptations, as well as studying the trade-offs for NLG.

\rparagraph{Contributions}
\textbf{1)} Related to Q1, we conduct a comprehensive analysis of ICL versus SFT and SIT paradigms in the context of multilingual few-shot adaptation, with the focus on multiple practical angles and cost. Our analyses show that not only the SFT and SIT approaches with smaller models lead to improved task performance but also they remain more data-, computation-, inference-effective than ICL with general-purpose LLMs. \textbf{2)} Related to Q2, we investigate the effectiveness of target language adaptation, adopted from the work on mPLMs, for ICL with LLMs. The main finding is that language adaptation leads to superficially improved generation capabilities in the target language with only limited improvements on the actual tasks, calling for further research that will mitigate the large language gap in LLM development and deployment between English and other languages.

%\ivan{We could add reflections on cross-lingual transfer and PEFT perhaps also in the "Limitations" section, as we're not doing cross-lingual transfer but rather direct few-shot learning in the target language. It is not even apparent how to marry the concept of cross-lingual transfer with ICL?}
%\ivan{SIT is never mentioned in the intro? Do we need to mention it?}

\section{Related Work}
\sparagraphnodot{Instruction-Tuning LLMs} aims to increase their cross-task generalisation capabilities. Instruction tuning is in essence an SFT technique where the input includes textual description of the task, demonstrations and user input queries while the output is the desirable model output for a given task in text form. Through inclusion of task descriptions into the input, at inference time the model becomes capable of completing tasks unseen during training when provided with task description \cite[interalia]{sanh2022multitask-t0, chung2022-flan}. Instruction tuning has become a standard approach to turn an LLM into a model with general capabilities to perform any task, given the instructions, off-the-shelf \cite{wei2021finetuned, mishra-etal-2022-cross}. 

\rparagraph{Extending LLMs to Other Languages} Although there is a growing trend to make NLP systems more linguistically inclusive \cite{bender2011achieving, doddapaneni2021primer}, widely used generative LLMs remain predominantly English. For instance, pretraining data of LLaMA-2 and PaLM consists of 90\% and 82\% English text, respectively \cite{touvron2023llama2, sitaram2023everything}, which substantially hinders their capabilities in languages other than English~\citep{ojo2023good}. In an attempt to equate the models' performance across languages, there is an increasing interest in extending their multilingual capabilities. A wide range of techniques including continued pretraining \citep{cui2023efficient}, using self-instruction \cite{wei2023polylm} or vocabulary extension \cite{zhao2024llama} have been applied. Due to wide adoption of ICL, another line of work focuses on improving cross-lingual instruction following capabilities via parameter-efficient multilingual instruction tuning \cite{li2023bactrian}, multilingual pretraining \cite{shliazhko2022mgpt} and injection of several multilingual examples in fine-tuning \cite{shaham2024multilingual}. The methods show gains in various aspects of model's target language generation capabilities, while providing no systematic empirical comparisons to prove that improved NLG necessarily correlates with stronger NLU performance via ICL. 
%neither of them manages to drastically improve the model's ICL capabilities.

These works provide initial insights into LLMs processing for languages other than English.  Interestingly, the success of mPLMs in cross-lingual transfer has always been attributed to their massively multilingual pretraining while, at first sight, LLMs seem to operate differently: they perform surprisingly well while having only a small percentage of multilingual text in their pretraining corpora \cite{blevins2022language-contamination}. At the same time, little to no work has studied multilingual performance of these models in direct comparison with standard mPLMs, and even more so the practical aspects such as memory requirements or latency.

\section{Preliminaries: On Learning Paradigms and Practical Aspects}
We analyse three established \textit{learning paradigms} for few-shot learning in monolingual and multilingual setups, which are compared across four \textit{practical aspects}: data cost, computational cost, inference cost and financial cost. We now outline each learning paradigm and practical aspect.

%% \ivan{Please be precise with terminology - are we really doing cross-lingual transfer here? No. I changed this accordingly.}

\subsection{Learning Paradigms}\label{subsec:learning_paradigms}
Let $\mathcal{D}={(x_1, y_1)}, ..., (x_N, y_N)$ denote a training dataset where $x_i$ is the model input, $y_i$ is the label annotation and $N$ is the number of training examples, and let $\mathcal{M}$ refer to a pretrained language model (LLM or mPLM).

\rparagraph{Supervised Fine-Tuning (\textsc{SFT})} $\mathcal{M}$ is adapted to a task or a language (or both) by fine-tuning its parameters on $\mathcal{D}$ and minimising a loss function. Note that here we use SFT in its narrower sense, to refer to `standard` fine-tuning where an encoder-based model (such as mBERT) or encoder-decoder model (e.g., mT5) is tuned directly for the target task \citep{devlin-etal-2019-bert, wei2022finetuned-sit-iclr}. At inference, the fine-tuned model $\mathcal{M}'$ is then used.    

\rparagraph{In-Context Learning (\textsc{ICL})} Unlike with \textsc{SFT}, the parameters of $\mathcal{M}$ stay fixed and the model is treated as a `black box'. ICL relies on generative capabilities of general-purpose LLMs \citep{brown2020-in-context-learning,han-etal-2023-understanding}. The model is adapted to a task by conditioning it on task instructions and in-context examples (\textit{demonstrations}). Each demonstration included into a prompt consists of an input $x$ and ground-truth annotated label $y$. In other words, the demonstrations are an alternative way to use the data available in $\mathcal{D}$. Then, $\mathcal{M}$ is expected to generate the label for the test input usually included at the end of the prompt. While in SFT the model parameters are adapted to a target task, with ICL the model is expected to learn the task by analogy, via the provided task description combined with demonstrations.

%predict output $y$ relying only on: i) the model's inductive bias to complete any task described in the instructions, and ii) learning by analogy from demonstrations. 

\rparagraph{Supervised Instruction-style Tuning (SIT)} To unlock full potential of ICL, sufficiently large language models need to be used~\cite{Wei:2022emergent}, drastically raising the computational overhead at inference in comparison with SFT. SIT thus presents the middle ground between the two. Here, one fine-tunes small(er) instruction-based models to specific tasks. While SFT fine-tunes the model directly on annotated data  $\mathcal{D}$, SIT extends each input in $\mathcal{D}$ with task-specific instructions leveraging model's instruction-following capabilities \cite{wei2022finetuned-sit-iclr} obtained during pretraining.  SIT typically does not include demonstrations into input, although including them there is also possible~\cite{min-etal-2022-metaicl,chen-etal-2022-meta}, typically with small to negligible performance gains in few-shot setups but increased computational cost~\cite{chengzu2023}. For simplicity, we experiment only with the SIT variant \textit{without} any demonstrations.

%While  SFT trains the model directly on inputs and their golden labels, SIT adds task instruction to the inputs, facilitating the use of    Note that the inputs here include only the task description (without any demonstrations, unlike in ICL). The models are tuned in sequence-to-sequence fashion and are tasked to generate the labels as text. 

\subsection{Practical Aspects}
We consider practical costs of the \textit{`full cycle`} of model development -- from data collection costs to inference cost, and aim to associate those costs with the learning paradigms described in \S\ref{subsec:learning_paradigms}. 

\rparagraph{Data Cost} One key limiting factor for the model adaptation to new task-language (or even finer-grained task-language-domain) combinations is the costly and complex data collection process, especially for low-resource languages and specialised domains. Therefore, it is crucial to develop methods which can efficiently generalise from a small number of annotated examples. In \S\ref{sec:results_main}, we analyse this data cost, that is, sample efficiency as the relationship between the number of training examples and task performance.

\rparagraph{Computational Cost} The memory requirements of LLMs keep growing proportionally to the number of their parameters. Deploying such a model to the users means that one needs to have access to and support costly infrastructure with large vRAM \cite{aminabadi2022deepspeed,alizadeh2023llm}. Here, we analyse the memory requirements of each learning paradigm both for model storage and training, where applicable, and how they correlate with the target task performance. 

\rparagraph{Inference Cost} Latency, or time needed by the model to complete the prediction {\cite[Chapter~1]{huyen2022designingmlmsys-latency}},
%\ivan{Do we have a reference for this definition?} 
has the largest impact on user-facing applications such as task-oriented dialogue. To make the system usable, it is critical to strike a balance between strong performance and low latency. We thus analyse the inference cost in two ways as: \textbf{1)} wall-clock inference time, aiming to directly approximate (relative) latency of different models; and \textbf{2)} inference FLOPs, a hardware-independent metric to compare inference complexity. 

\rparagraph{Financial Cost} Each of the aspects above contributes to the overall cost of each model's life cycle. As financial resources are usually limited, we also aim to (roughly) estimate the overall financial expenditure needed for each learning paradigm, including data collection, GPU and inference costs.    
%\footnote{We critically need some citations in this section - both for learning paradigms as well supporting evidence for all the cost aspects we talk about}

% \ivan{The paper critically requires a standalone section that will properly define all those practical aspects beyond performance (listed in Introduction after Q1) and how we'll measure them and why they're practically important. This is the key to the paper.}
% \ivan{Another subsection of that section should focus on quickly defining learning paradigms, but before talking about the actual models used.}
\section{Experimental Setup}
\label{sec:setup_sft_vs_icl}
We focus on the comparison between SFT, SIT and ICL in few-shot multilingual and cross-lingual setups, aiming to make the comparison as fair as possible across languages, learning paradigms and models, and targeting the following setups:

\rrparagraph{In-Language Generalisation} We evaluate the model's ability to generalise on new examples in the same language in which fine-tuning examples or demonstrations were provided to the model. %either the model was finetuned on or seen in demonstrations.

\rrparagraph{Cross-Language Generalisation} We use a model trained in one language to perform the task in another one, where the transfer typically proceeds from a high-resource language to a low-resource one. In our experiments, we assume the typical transfer direction with English as the (high-resource) source language.

\rrparagraph{In-Domain and Cross-Domain Generalisation} For many NLU tasks (e.g., for task-oriented dialogue) it is common to consider transferring the systems between different domains, e.g., from \textit{flight booking} to the \textit{restaurant booking} domain. If a model can be transferred across domains, it means that it has in-depth understanding of the classes used in the respective domain definitions/ontologies. 

\subsection{Evaluation Tasks and Datasets} 
\label{ss:tasks}
%We conduct comparison three classification tasks: a) a token classification task, value extraction (VE); b) a sentence classification task, intent detection (ID); and c) between-sentence classification task, natural language inference (NLI).  

%\setlength{\tabcolsep}{4.5pt}

%% \textsc{classes} correspond to the total number of classes per dataset. 
%% For Multi3NLU++ the split of per-domain classes is 26:22:14 and 10:3:4 for general:\textsc{banking}:\textsc{hotels} for \textsc{id} and \textsc{ve}, respectively.

The main focus of the analyses, revolving around Q1 and Q2 from \S\ref{sec:intro}, is on NLU tasks for task-oriented dialogue as one widely used and established practical application of NLP technology, due to multiple reasons. \textbf{1)} Dialogue is a user-facing application where computational and memory requirements, data collection cost, inference latency and other practical concerns of the model development cycle are of ultimate importance. \textbf{2)} Dialogue NLU tasks provide well-defined ontologies and evaluation setups, with evaluation benchmarks that comprise comparable and semantically aligned training and test data across multiple languages, including high- and low-resource ones \cite{moghe-etal-2023-multi3nlu, hu2023multi3woz}, and multiple domains. \textbf{3)} In contrast to standard `non-dialogue' NLU tasks, dialogue NLU datasets are unlikely to have been seen and `absorbed` by LLMs during their pretraining, which avoids test data leakage \citep{balloccu2024leak, sainz2023nlp}. 

%\ivan{Remember to add citations to CITE in the whole paragraph}

Dialogue-oriented evaluation is conducted on the tasks of intent detection (ID) and value extraction (VE). \textsc{ID} aims to classify user's utterance into a set of intent classes predefined in the domain ontology. The aim of \textsc{VE} is to identify the presence of ontology-related domain-specific slot-value pairs in a given sentence. Here, we use the Multi3NLU++ dataset \cite{moghe-etal-2023-multi3nlu}, covering English \citep{casanueva-etal-2022-nluplusplus} and the following 4 languages:  Amharic (\textsc{am}), Marathi (\textsc{mr}), Spanish (\textsc{es}) and Turkish (\textsc{tr}). The dataset also spans two different domains: \textsc{banking} and \textsc{hotels} with a partial overlap in intent classes and slots. For both tasks we report micro-F1 scores.\footnote{We also note that \textbf{1)} each utterance in Multi3NLU++ may have multiple intents; ID is thus a multi-label classification task. \textbf{2)} Further, for VE, we consider the slot value as correctly labelled only if it exactly matches the gold value. Finally, \textbf{3)} as in prior work \citep{casanueva-etal-2022-nluplusplus, moghe-etal-2023-multi3nlu}, the cross-domain performance for the two tasks is evaluated only on the intents and slots shared across domains. We refer to the original Multi3NLU++ work for further details.}

To verify that our findings extend beyond only dialogue-related NLU tasks, we also evaluate on the standard NLI task with XNLI~\citep{conneau-etal-2018-xnli} which provides training and evaluation data in 14 languages, while we focus on a subset of 3: \textsc{es}, \textsc{tr}, and Russian (\textsc{ru}), and report accuracy as the evaluation metric. Additional information on the evaluation datasets is provided in Table~\ref{tab:dataset_stats}.

%Appendix\ref{app:dataset_stats}.

\begin{table}[!t]
\def\arraystretch{0.71}
{\fontsize{6.9}{7.1}\selectfont
%\footnotesize
\resizebox{\linewidth}{!}{%
\begin{tabular}{ll llll}
%\begin{tabularx}{0.99\linewidth}{ll llll}
\toprule
\multirow{1}{*}{\em Dataset} &  & \textsc{langs} & \textsc{\# test ex.} & \textsc{\# classes}                              \\ \midrule
\multirow{2}{*}{\textit{Multi3NLU++}}  & \textsc{id}                     &    \multirow{3}{*}{\shortstack[c]{\textsc{am}, \textsc{en}, \textsc{es}, \textsc{mr}, \textsc{tr}}}     & \multirow{2}{*}{300} & 62 \\
                     &     \multirow{2}{*}{\textsc{ve}}                 &     &  & \multirow{2}{*}{17}                 \\  
                     &                      &     &  &                 \\                      \cmidrule{2-5}
\multirow{2}{*}{\textit{XNLI}}  & \multirow{2}{*}{\textsc{nli}}                     &    \multirow{2}{*}{\shortstack[c]{\textsc{en}, \textsc{ru}, \textsc{tr}, \textsc{es}}}   & 
\multirow{2}{*}{5,010} & \multirow{2}{*}{3}   \\
                     &                      &     &  &                 \\
 \bottomrule
%\end{tabularx}
\end{tabular}
}
}%
\vspace{-1mm}
\caption{Summary of evaluation datasets and tasks.} \label{tab:dataset_stats}
\vspace{-2mm}
\end{table}

\rparagraph{Cross-Language Parallel Few-Shot Setup} 
To ensure fair comparisons of all learning paradigms across languages, we make use of the multi-parallel nature of the datasets we use. For each \textit{language-domain} combination in Multi3NLU++ (or just \textit{language} in XNLI) we sample 300 test examples.\footnote{We conduct the sampling step due to a large number of experiments run; preliminary experiments with full test sets indicated exactly the same relative trends, but with much increased computational cost and time overheads. While sampling, we ensured that each intent and slot in the domain ontology occurred at least twice in the test set} We also randomly sample training sets consisting of \{30, 50, 100, 500, 1,000\} examples which are kept exactly the same across all languages to ensure the content in training data does not coincidentally favour one of the languages.\footnote{To ensure reproducibility, unique ids of the examples in the training and test splits will be made publicly available.} 

\rparagraph{SFT Evaluation} 
We use two standard models, \textit{XLM-R-Base}~\cite{conneau-etal-2020-unsupervised}  and \textit{LaBSE} \cite{feng-etal-2022-labse}. LaBSE is a sentence encoder model where, following prior work~\cite{moghe-etal-2023-multi3nlu}, we train only task-specific classifiers on top of the fixed encoder. We refer to this approach, applied only to sentence-level tasks (ID and NLI), as \textit{LaBSE+CL}.\footnote{For NLI as a single-label classification task, the \textit{softmax} output activation function is used. In contrast, for ID we use \textit{sigmoid} and consider all intents where the \textit{sigmoid} activation is larger than a predefined threshold value $\theta$. Following prior work, we set $\theta = 0.3$.} 
%% Details of training hyperparameters are provided in Appendix \ref{app:hyperparameter_values}

\rparagraph{ICL Evaluation} 
We evaluate the following models: Flan-T5-XL, mT0-XL, LLaMA-2-7B and GPT-3.5.\footnote{We use \textit{GPT-3.5-turbo-instruct} due to its instruction-following capabilities proven in prior work~\citep{ye2023comprehensive}} Flan-T5-XL (3B parameters; \citeauthor{chung2022-flan}, \citeyear{chung2022-flan}), mT0-XL (3.7B; \citeauthor{muennighoff2022-mt0}, \citeyear{muennighoff2022-mt0}) and GPT-3.5 \citep{achiam2023gpt4technical} are massively instruction-tuned models. While Flan-T5 was pretrained mostly in English and several high-resource languages, mT0-XL offers a more comprehensive and balanced multilingual pretraining set. %Although LLaMa-2 was not specifically instruction-tuned, it enables instruction following via chat. 

The inputs for ICL were designed in a cross-lingual manner, where the task descriptions and context were in English while the few-shot examples and the sentence to be analysed were provided in the target language. This follows the recommendations from prior work where it was empirically verified that English instructions led to stronger results than in-language instructions \cite{shi2022language-multilingual-cot,lin-etal-2022-shot}. For each task we design the instructions (i) to match the instructions in pretraining as closely as possible, while (ii) yielding reasonable output when tested on several validation examples.\footnote{For reproducibility, we will share the full instructions templates for all languages and tasks.} Note that, given a fixed input context of each model, the number of demonstrations to be used for ICL is limited: for all the models in our comparison it is less than 30, and 30 is the lowest amount of training samples we use in SFT.

%%  Appendix \ref{sec:instructions_for_tasks} presents the texts of the instructions used in the experiments.

\rparagraph{SIT Evaluation} Here, we include individual per-class questions into instructions: this design (i) was previously shown to result in much improved SIT performance \citep{Fuisz2022improvedve-qa,razumovskaia2023sqatin},  while (ii) it also fits into the model input context for tasks with a large number of classes. We rely on the same instructions as with ICL. We experiment with two models: (i) (mostly English pretrained) \textit{Flan-T5-Base} (250M parameters) and multilingually oriented \textit{mT0-Base} (580M).\footnote{{SFT and SIT training hyperparameters are in Appendix~\ref{app:hyperparameter_values}.}}

%% (IV, removed, said before)
%% With the SIT paradigm, instruction-tuned models are fine-tuned in a supervised sequence-to-sequence fashion, where the input is a combination of task description and a training example and the text output is the label. 

\section{(Q1) Results and Discussion: Learning Paradigms and Practical Aspects}
\label{sec:results_main}
We first delve into comparisons revolving around Q1 (\S\ref{sec:intro}). The main results across different setups, models, training data sizes and learning paradigms are summarised in Figure~\ref{fig:id_few_shot_results}, while full (numerical) results are provided in Appendix~\ref{app:full_experimental_results}. 
We now zoom into discussions originating from the results.
\captionsetup[subfigure]{oneside,margin={-0cm,0cm},skip=-1pt}
\begin{figure*}[!t]
\centering
\begin{subfigure}[b]{0.95\textwidth}
   \includegraphics[width=1\linewidth]{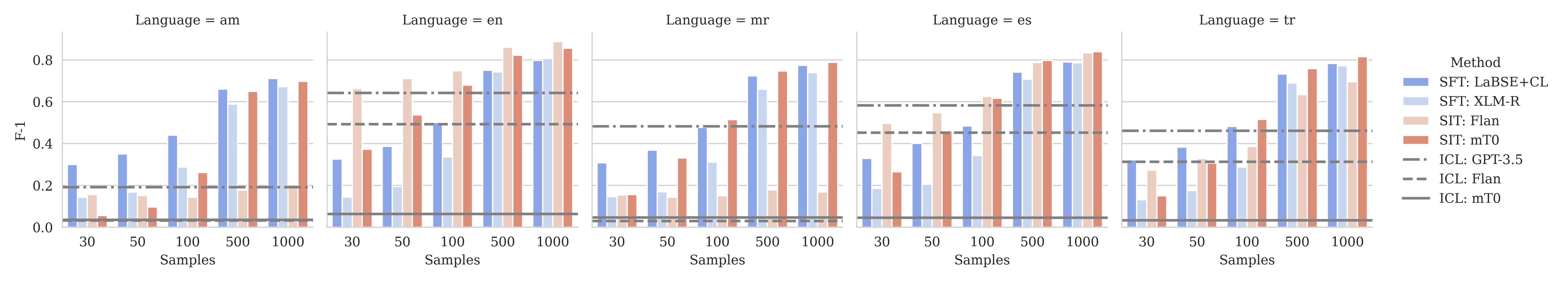}
   \caption{\textsc{id}: In-Language In-Domain}
   \label{fig:ID_indomain_inlanguage} 
\end{subfigure}

\begin{subfigure}[b]{0.95\textwidth}
   \includegraphics[width=1\linewidth]{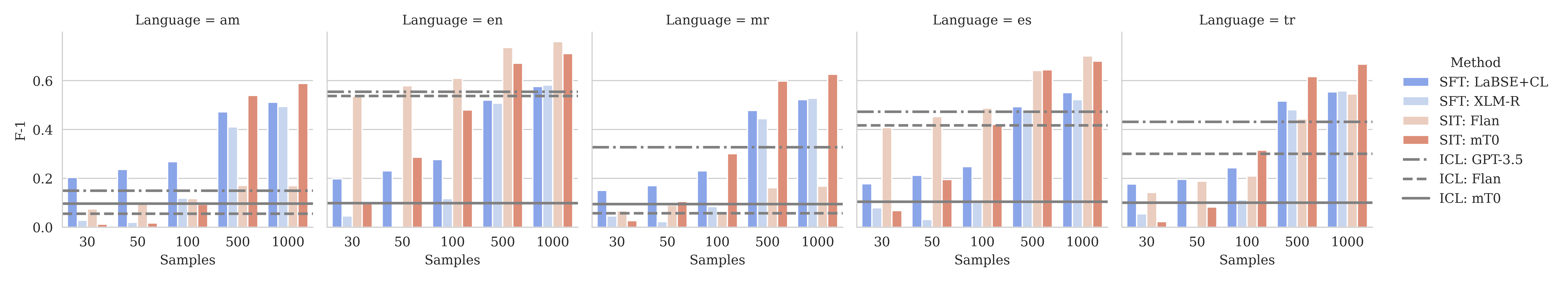}
   \caption{\textsc{id}: In-Language Cross-Domain}
   \label{fig:ID_crossdomain_inlanguage}
\end{subfigure}

\begin{subfigure}[b]{0.95\textwidth}
   \includegraphics[width=1\linewidth]{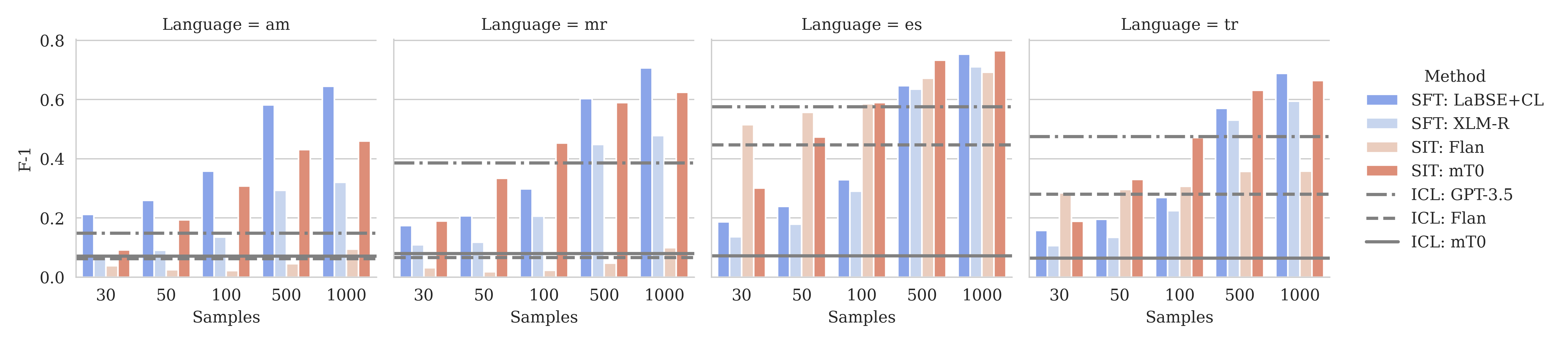}
   \caption{\textsc{id}: Cross-Lingual In-Domain}
   \label{fig:ID_indomain_crosslingual}  
\end{subfigure}

\begin{subfigure}[b]{0.95\textwidth}
   \includegraphics[width=1\linewidth]{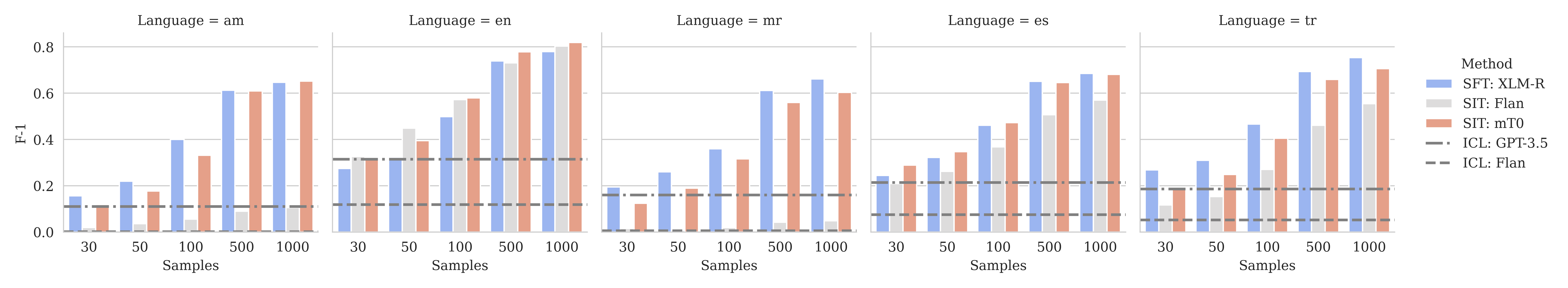}
   \caption{\textsc{ve}: In-Language In-Domain}
   \label{fig:SL_indomain_inlingual}  
\end{subfigure}

\begin{subfigure}[b]{0.95\textwidth}
   \includegraphics[width=1\linewidth]{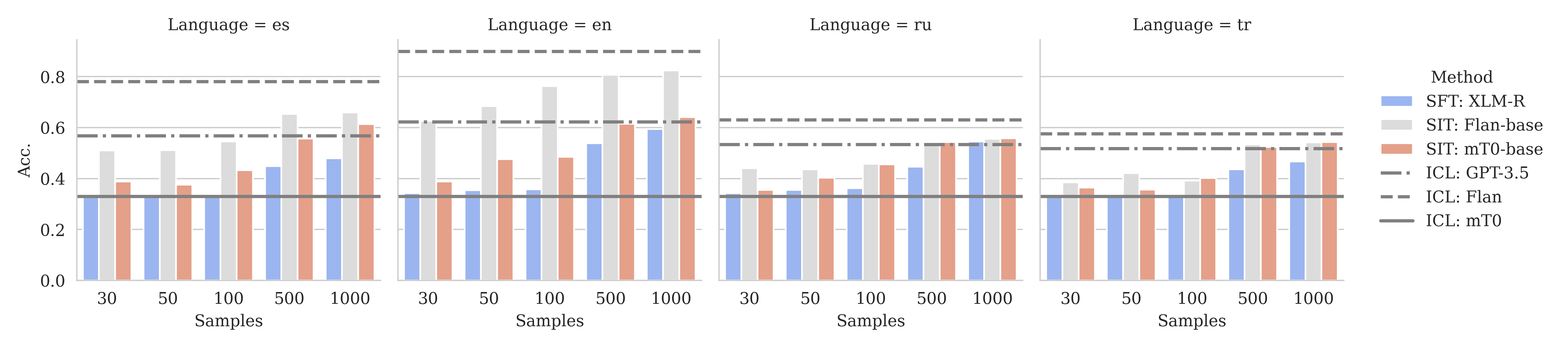}
   \caption{\textsc{xnli}}
   \label{fig:xnli_results}  
\end{subfigure}
\vspace{-1mm}
\caption{Intent detection, value extraction and NLI results for the six languages in our evaluation. This performance is in line with other prior work~\cite{hu-etal-2023-systematic}. We exclude LLaMa-2 results as its performance was 0.0 across all tasks. Results for \textsc{ve} in other setups are provided in Appendix~\ref{app:other_ve_results}.} \label{fig:id_few_shot_results}
\vspace{-2mm}
\end{figure*} 

\rparagraph{Data Efficiency} 
One of the core reasons to use ICL is its inherent data/sample efficiency. Comparing the supervised methods against ICL, we observe that the former reach or overcome the performance of ICL with all tested open-source models (Flan-T5-XL, mT0-XL and LLaMA-2-7B), even when fine-tuned with mere 30 in-task examples, while they outperform GPT-3.5 with 50 or 100 in-task examples. These findings hold across all evaluation tasks and setups. 

At the same time, the results also reveal several key differences between tasks and languages. Comparing the trends for \textsc{id} and \textsc{ve} (cf., Figures \ref{fig:ID_indomain_inlanguage} and \ref{fig:SL_indomain_inlingual}): for sentence classification tasks where the outputs are language-independent, the improvements of SFT over ICL are less pronounced. For instance, the gains over GPT-3.5 with 100 training examples for \textsc{mr} and \textsc{es} are 3.15 and 4.17 F1 points,\footnote{The cited numbers are for in-language in-domain setups; the trends are the same in the other setups.} respectively. In contrast, the gains over ICL for value extraction as a more language-specific task are very large, 19.8 and 25.28 for \textsc{mr} and \textsc{es}, respectively, when comparing the best-performing SFT method with ICL. Moreover, for \textsc{ve}, the gaps with ICL are considerable even with 30 training examples are used (3.4 and 7.5 $F_1$ points for \textsc{mr} and \textsc{es}).

 %Other interesting patterns emerge from comparing performance of SFT and ICL across languages.
 
 For \textsc{am} a supervised model surpasses GPT-3.5 performance already with 30 training examples, while for \textsc{es} 50 or 100 training examples are required, depending on the setup. For high-resource languages (\textsc{en, es}) SIT-based Flan-Base with 30 examples performs consistently better than ICL with GPT-3.5 ICL for \textsc{id} and \textsc{ve} across the setups. We hypothesise that high performance of SIT-based Flan is caused by i) its instruction following capabilities, and ii) large-scale English pretraining which is helpful for both few-shot in-language generalisation and cross-lingual transfer from English to Spanish, a linguistically close language.  For low-resource languages (\textsc{am, mr}) we notice a different tendency across the domain setups: LaBSE+CL and SIT-based mT0 show the highest performance for \textsc{id} and \textsc{ve}, respectively. This shows the importance of multilingual pretraining for the model to generalise to unseen or `less seen' languages.

%We speculate that high performance of Flan on high-resource languages might stem from: (i) potential data leakage of non-English high-resource language into pretraining \citep{blevins2022language-contamination}; b) linguistic similarity between English and other high-resource languages in evaluation. 

%The trend for low-resource languages (\textsc{am, mr}) is drastically different: LaBSE+CL demonstrates the highest performance in sentence-level tasks when tested in-domain while being on-par or outperformed by SIT-ed mT0 in other tasks and setups. This demonstrates that multilingual pretraining is essential for few- and zero-shot cross-lingual transfer to low-resource languages.\ivan{The part form 'we speculate' to the end of the paragraph should be rewritten: very imprecise overall}  

\rparagraph{In-Domain vs Cross-Domain Evaluation} 
The comparison in cross-domain setups consistently shows that SIT outperforms SFT and ICL (see Figure~\ref{fig:ID_crossdomain_inlanguage}), corroborating findings from prior work on English \cite{razumovskaia2023sqatin}. We speculate that the success of SIT in cross-domain setups stems from the model's ability to follow instructions obtained during pretraining and the ability to extract the class semantics from instructions obtained during fine-tuning. The best-performing instruction-tuned LLM, however, depends on the target language: on low-resource languages multilingually pretrained models such as mT0 perform consistently better than English-pretrained models such as Flan, while we observe reversed trends for high-resource languages (\textsc{es}).

\rparagraph{Cross-Lingual Zero-Shot Transfer} 
Figure~\ref{fig:ID_indomain_crosslingual} presents the results for zero-shot transfer in in-domain setups: performance across languages for all approaches is substantially lower than the performance in English. Further, as expected, performance on low-resource languages is considerably lower than on high resource languages. 

The results in Table~\ref{tab:icl_from_english_or_tgt} show that for ICL, unlike for SFT \cite{lauscher2020zero-to-hero}, providing the model with data examples in the target language does not always improve the final performance. Target language demonstrations seem to be helpful to the models which have strong instruction-following capabilities and are familiar with the target language (e.g., \textsc{es} performance of Flan and GPT-3.5). 

\rparagraph{Seen vs Unseen Tasks}
Figure~\ref{fig:xnli_results} demonstrates the effectiveness of ICL with Flan and GPT-3.5 for \textsc{xnli} as the `seen task',\footnote{XNLI is based on the English MultiNLI data~\cite{williams-etal-2018-broad}, which has been used for instruction-training of many LLMs~\cite{muennighoff2022-mt0, chung2022-flan}.} with different patterns observed for the tasks with unseen data (i.e., \textsc{id} and \textsc{ve}). This discrepancy is especially pronounced for high-resource languages. %We assume that this is due to the familiarity of the task to the model. 

\setlength{\tabcolsep}{4.5pt}
\begin{table}[!t]
\def\arraystretch{0.81}
{\fontsize{6.9}{7.1}\selectfont
%\footnotesize
%\resizebox{\linewidth}{!}{%
\begin{tabularx}{0.99\linewidth}{ll XXXX}
\toprule
\rowcolor{Gray}
\multirow{1}{*}{\em Model} &  & \textsc{am} & \textsc{mr} & \textsc{es} & \textsc{tr}                               \\ \cmidrule{2-6}
\multirow{2}{*}{\textit{GPT-3.5}}  & ICL$_{t}$                &        19.19 & 48.28 & 63.25 & 59.27            \\
                     &  ICL$_{en}$                    &    14.87 & 38.67 & 57.64 & 47.50                  \\  \cmidrule{2-6}
\multirow{2}{*}{\textit{Flan}}  & ICL$_{t}$                   &     3.28  & 3.02  & 45.26 & 31.36   \\
                    &  ICL$_{en}$                    &      6.33  & 6.68  & 44.70 & 28.04                                 \\ \cmidrule{2-6}
\multirow{2}{*}{\textit{mT0}}  & ICL$_{t}$                  &      3.61  & 4.71  & 4.60  & 3.36          \\
                    &  ICL$_{en}$                    &         7.19  & 7.99  & 7.23  & 6.45            \\
 \bottomrule
\end{tabularx}
}
%}
\vspace{-1mm}
\caption{ICL results on the \textsc{id} task with English (ICL$_{en}$) or target language (ICL$_{t}$) demonstrations.} \label{tab:icl_from_english_or_tgt}
\vspace{-2.5mm}
\end{table}

\rparagraph{SIT vs ICL}
In general, the results indicate that SIT consistently leads to better results than ICL in few-shot setups. Smaller SIT-based models can even outperform ICL with GPT3.5 when 100+ task examples are available. Due to its sample efficiency and strong performance in cross-domain and cross-lingual setups, SIT also mitigates the issue of using a separate model for each \textit{task-language} or \textit{task-language-domain} combination. 

\begin{table}[!t]
\def\arraystretch{0.81}
%{\fontsize{6.9}{7.1}\selectfont
%\footnotesize
\resizebox{\linewidth}{!}{%
%\begin{tabularx}{0.999\linewidth}{l XXXX}
\begin{tabular}{l cccc}
\toprule
 \multirow{2}{*}{{\bf Paradigm: Model}} & \multicolumn{2}{c}{\textsc{memory} cost} & \multicolumn{2}{c}{\textsc{inference} cost} \\ \cmidrule(lr){2-3} \cmidrule(lr){4-5}
                      & \textit{Max (GB)}    & \textit{Storage (GB)}   & \textit{Time (s)}    & \textit{FLOPs} ($10^9$)  \\ \cmidrule(lr){2-3} \cmidrule(lr){4-5}
SFT: LaBSE+CL  & 1.80                            & 1.80                             & 0.004                          & 2.05                      \\
SFT: XLM-R     & 4.14                            & 1.04                             & 0.004                          & 11.18                     \\
SIT: Flan-Base & 3.32                            & 0.85                             & 0.059                          & 23.23                     \\
SIT: mT0-Base  & 5.82                            & 1.45                             & 0.081                          & 20.44                     \\
ICL: Flan-XL   & 10.37                           & 10.37                            & 1.58                           & 39.03                     \\
ICL: mT0-XL    & 12.03                           & 12.03                            & 1.20                           & 55.00                     \\
ICL: GPT-3.5   & -                               & -                                & 2.78                           & -                         \\ \bottomrule
\end{tabular}
}%
\vspace{-1mm}
\caption{Memory and inference costs of SFT, SIT and ICL, measured on the \textsc{id} test dataset. \textsc{memory} \textit{Max} is the peak fine-tuning or storage memory cost of each approach. \textsc{memory} \textit{Storage} refers to storage requirements per models. For all models but GPT-3.5 the measurements were conducted on a single RTX-3090 GPU. For GPT-3.5, we report average response time per example. }\label{tab:costs_methods}
\vspace{-1.5mm}
\end{table}

\subsection{Analyses of Practical Costs}
Given that the results above indicate that the two supervised paradigms (SFT and SIT) substantially outscore ICL in terms of task performance in general, we now focus on comparing them in terms of practical aspects. The summary is presented in Figure~\ref{fig:comparison_practicalities} (see \S\ref{sec:intro}) and Table~\ref{tab:costs_methods}.%, and we discuss it in what follows.

\rparagraph{Computational (Memory) Cost}    
Besides improved task performance, another advantage of SFT and SIT is that the underlying high-performing models are much smaller and thus have lower memory requirements. The largest memory cost for ICL is storing the model’s parameters at inference time while for SFT and SIT it is the memory requirements during fine-tuning. We rely on \href{https://huggingface.co/docs/accelerate/main/en/usage_guides/model_size_estimator}{HuggingFace Memory Calculator} to establish vRAM needed for training and inference of every paradigm. We measure the memory requirements in full precision and using the AdamW optimiser~\cite{adamw}, when applicable.\footnote{Closed-source GPT-3.5 is excluded from the comparison.} The results indicate that models used for ICL have more than $2\times$ higher memory needs than mT0 and Flan-T5-base used for SIT in our experiments. Another angle to memory requirements is the storage cost, i.e., how much memory is needed to store a given model (`as is' for ICL and after fine-tuning for SFT and SIT). Table \ref{tab:costs_methods} suggests that storage cost for models used for ICL is at least $4\times$ higher than the models used in SIT and SFT.

%This means that we could store around 4 models SIT-ed for different tasks in the same space as one mT0-XL or Flan-XL checkpoint, with superior performance across domains and languages. 

%Using larger models for ICL means not only higher memory requirements, but also inference cost, which is especially important for user-facing applications such as task-oriented dialogue, as the user experience is highly dependent on latency. 
\rparagraph{Inference Cost}
Beyond average wall-clock inference time per test example. we also report the number of FLOPs measured using \href{https://github.com/facebookresearch/fvcore}{fvcore}, {also averaged per test example}. %\ivan{Is this also the average per example?} 
As expected, the inference cost scales with the size of the underlying model, with inference time of GPT-3.5 being more than 3x higher than that of SIT-ed models, {and inference FLOPs of open-source ICL models being $2.5\times$ higher than for their smaller SIT-ed counterparts.
%\ivan{I don't understand what is lower than what here? You're not reporting FLOPs for GPT3.5 in the table} 
While SFT methods demonstrate even higher inference efficiency, we observe that SIT has the best trade-off between inference cost and performance. 

\rparagraph{Financial Cost} Having demonstrated considerably higher inference costs of ICL, we also consider overall economic costs required for SFT, SIT, and ICL. Target language data annotation accounts for the largest expenditure in the process. We calculate the annotation cost based on \citet{moghe-etal-2023-multi3nlu}. ICL consumes up to 30 annotated examples, with total costs of \pounds 15.9 and \pounds 18.6 for high-resource and low-resource languages, respectively, where the annotations get obtained both for \textsc{id} and \textsc{ve}. In the \textsc{ve} task, SIT-based methods reach or surpass the ICL performance of the strongest model (GPT-3.5) already with 20 extra examples (i.e., with 50+ training examples for supervised learning), which only adds \pounds 11 or \pounds 10 to the overall cost for low- and high-resource languages, respectively. 

Given the larger inference time and computational costs of the ICL, the total ongoing costs are likely to be larger than the one-time additional annotation budget. To put the numbers in context, the inference cost of 300 test examples with GPT-3.5 is between \pounds 3 and \pounds 4 for high- and low-resource languages, respectively. Put simply, the actual cost balance should take into account also the tentative number of inference calls.

Further, while increasing the input context length of LLMs is an active research area~\cite[\textit{among others}]{alibi,Rubin:2023arxiv}, many models relying on the ICL paradigm are still constrained by context length, and there is evidence that ICL performance even gets quickly saturated with the addition of extra in-context examples~\cite{chen-etal-2023-many,chengzu2023} and that the long context is not leveraged adequately~\cite{lostinthemiddle}. On the contrary, unlike with ICL, our experiments demonstrate that performance of SFT and SIT improves with more annotated examples (both in-language and cross-lingually, see Figure~\ref{fig:id_few_shot_results}). Data annotation of 100 training examples raises the annotation cost by an average of \pounds 37.5 while {increasing the \textsc{id} and \textsc{ve} performance by an average of 15 F-1 points over ICL with GPT-3.5.} 
%providing gains of over 30 F1 points .\ivan{Gains over what? GPT3.5 or their ICL variants or something else? What task?}

\section{(Q2) Results and Discussion: Target Language Adaptation of LLMs}
\label{sec:adaptation}
\S\ref{sec:results_main} indicates that ICL is consistently inferior to the two supervised learning paradigms, SFT and SIT, not only in terms of task performance but also concerning computational and inference costs. At the same time, ICL relies on a single model and is thus appealing when extending a system to a large number of {language-task} combinations. 

Prior work on `decoder-only' LLMs  demonstrated the effectiveness of parameter-efficient fine-tuning (PEFT) to improve their cross-task generalisation capabilities \cite{page2024crosstask-peft}, whereas PEFT is a standard approach for cross-lingual adaptation of `encoder-only' and `encoder-decoder' models such as XLM-R or mT5 \cite{conneau-etal-2020-unsupervised, xue-etal-2021-mt5}. In this work, we explore whether such language-specific PEFT-style adaptation can improve ICL and generation capabilities of LLMs in languages other than English. We focus on LLaMA-2-7B as our base model, as: (i) it is a `decoder-only' model that (ii) has been trained as the `English-first' model, with almost 90\% of its pretraining data in English; and (iii) it displayed the lowest performance in our experiments in \S\ref{sec:setup_sft_vs_icl} while being the largest model in our evaluation. %in terms of its parameter budget. 

%\genie{I thought about it a bit more -- I think I would leave the Bactrian-X out of the main paper. It seems like it would only open a set of tangential questions and comparisons like: multisource vs only target language adapter training; general language adaptation vs instruction tuning in multiple languages... what do you think?}

\rparagraph{Language Adaptation Setup} We use QLoRA \cite{dettmers2023qlora} as a standard PEFT-based language adaptation technique. QLoRA performs two modifications to the base LLM. The model is first quantised to reduce the memory requirements and then a low-rank adapter \cite{hu2022lora} is trained on top of the quantised model. In our experiments the adapter is tuned on the target language data, aiming to boost the target language capabilities of the underlying LLM.

For the adaptation experiments, we focus on three languages: Spanish, Turkish and Marathi.
%\ivan{Add a footnote discussing that the actual adaptation is quite expensive - how many GPU hours per one adaptation?} 
The adapter for each language is trained on the respective portion of mC4 \cite{xue-etal-2021-mt5}. Hyper-parameters are set following~\citet{dettmers2023qlora}, with exact details available in Appendix~\ref{app:qlora_hyperparameters}.\footnote{{Training each QLoRA adapter requires over 24 GPU-h.}}

\subsection{Generation after Language Adaptation?}
First, we assess whether target language adaptation boosts generation capabilities of the LLM in the target language. To this end, we use the Bactrian-X dataset \cite{li2023bactrian}, a multilingual instruction dataset containing parallel instruction-response pairs in 52 languages. For our evaluation, we  use a subset of 100 randomly sampled examples ensuring the same parallel examples across the three languages in our evaluation (\textsc{es}, \textsc{tr}, \textsc{mr}). 

\rparagraphnodot{Generation Evaluation: True or Superficial Improvements?} We focus on the three aspects of generation capabilities: (i) whether the model outputs text in the same language as expected by the input (i.e., \textit{I/O language agreement}, similarly to \citeauthor{kew2023turning}, \citeyear{kew2023turning}); (ii) \textit{naturalness} of the generated text; (iii) \textit{lexical overlap} between golden responses and generation outputs. I/O language agreement involves doing automated language identification of the generated text and establishing whether it corresponds to the input text. For this purpose, we use the current state-of-the-art language identification model, GlotLID-500 \citep{kargaran-etal-2023-glotlid}. We evaluate naturalness via MAUVE~\cite{pillutla-etal:mauve:neurips2021} which measures the distributional gap between human written and generated texts. For lexical overlap, we report ROUGE~\citep{lin-2004-rouge} and BLEU~\citep{Papineni-2002-bleu}. 

Figure~\ref{fig:generation_results_qlora} shows consistent gains of generation capabilities over the three evaluation aspects after target language adaptation, with especially large improvements for Marathi as the lowest-resource language. The I/O agreement scores suggest that through language adaptation LLM's abilities to generate text in the target languages are reinforced.  

However, those standard metrics still do not fully capture the potential usefulness of generated output and (improved) generation capabilities. We thus also conduct human-based evaluation for Spanish across the following two axes: \textit{naturalness} and \textit{usefulness}. Each output is evaluated on a simple 3-point Likert-like scale.\footnote{Annotation instructions are provided {in Appendix~\ref{app:annotation_instructions}}.} %\ivan{Remember to add this appendix with instructions for humans} 
Interestingly, the average naturalness score raises from 1.4 to 2.2 after language adaptation while usefulness only increases from 1.4 to 1.6. In practice, this means that even after language adaptation the model is still far from being useful for the target language speakers. This finding corroborates preliminary observations of \citet{kew2023turning} that the English-centric models can learn to generate text in a target language comparatively easily, but useful instruction-following capabilities still remain largely out of reach. Put simply, while generated text in the target language becomes more fluent, its coherence and relevance remain limited. 
%% \ivan{Do we have these numbers for English? Can we quickly compute them?} \genie{No, I don't have the numbers for English and also I haven't trained an adapter on English mC4.}

\begin{figure}[!t]
\centering
\includegraphics[width=0.95\linewidth]{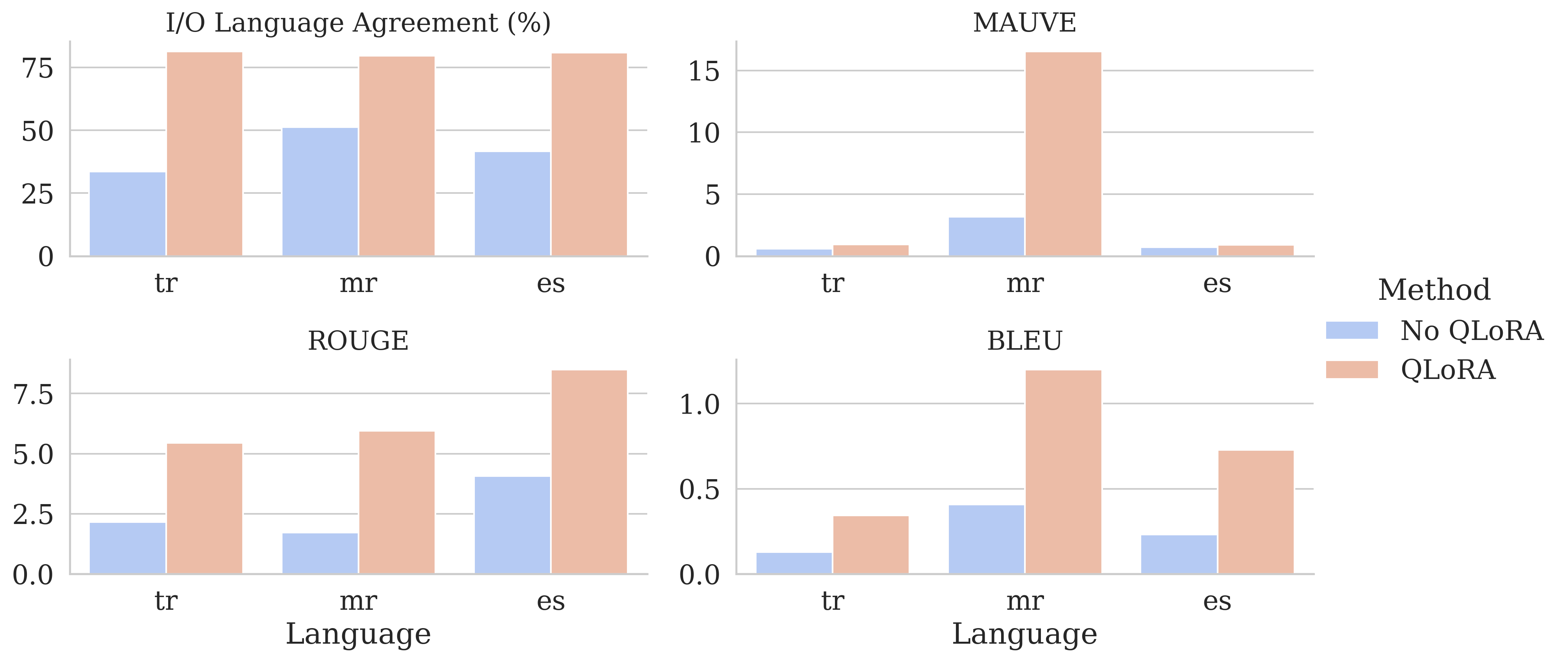}
\vspace{-2mm}
\caption{Generation evaluation after target language adaptation (LLaMA-2).}\label{fig:generation_results_qlora}
\vspace{-2.5mm}
\end{figure}

%\ivan{Let's break Figure 3 into the 2x2 grid rather than the current 1x4, it's not readable atm.}
%\ivan{Could you make subfigures in Figure 2 flatter, they will convey the same information in fewer space? Make them more 'rectangular'.}

\subsection{NLU after Language Adaptation?}
Given only superficial improvements in generation capabilities, we now assess whether the ICL capabilities improve for NLU tasks. For brevity, we focus on XNLI as the least complex NLU task. Even for XNLI, we observe only a negligible non-significant improvement from the average accuracy score of 30.5 to 30.7.\footnote{Per-language scores are provided in Appendix~\ref{app:xnli_results}. Similar relative trends have been observed in preliminary experiments on another, more complex NLU task: Belebele~\cite{Bandarkar:2023belebele}, where the results are on-par or lower than the random choice baseline, as well as in more complex NLU tasks from our evaluation in \S\ref{sec:results_main}.}  Performance is in fact below random (33\%), supporting the observations from \S\ref{sec:results_main} that resource-efficient ICL requires both multilingual pretraining and instruction tuning. From qualitative assessment of the outputs, we notice that the models struggle to follow the task description and instructions, and often do not adhere to output formatting requirements. 

\rparagraph{Massively Multilingually Adapted LLMs in NLU tasks} 
The results above suggest that \textit{direct} target language adaptation of `English-first' LLMs such as LLaMA-2 does not yield any benefits to ICL performance in NLU tasks. Next, we study whether \textit{massively multilingual} adaptation of `English-first' models, as done in very recent work, can improve their ICL capabilities in different languages. We evaluate the MaLa-500 model \cite{lin2024mala500} which was adapted for 534 languages using the Glot-500-c dataset \citep{imanigooghari-etal-2023-glot500}, and is also based on LLaMA-2.\footnote{MaLA-500 was adapted using: (i) LoRA-based parameter adaptation; (ii) vocabulary extension to accommodate for languages that do not use the Latin script.} We focus on the \textsc{id} task to evaluate MaLa's ICL performance in the (easiest) in-language in-domain setup, with results summarised in Table~\ref{tab:icl_llama_vs_mala}.
%% (IV, return to CR to the footnote: " as their script were not covered by LLaMA-2's original tokenizer")
%\ivan{Remember to add translate-test results into the table (1 line of results)} 
They reveal that, while adaptation gives marginal improvements for ICL, the performance is still extremely low and lags substantially behind GPT-3.5 performance and SIT with mT0-Base {with 100 training examples. A comparison with \textit{translate-test} baseline shows that while translate-test benefits low-resource \textsc{am}, it is still outperformed by SIT on all other languages.} 
%\ivan{How many training examples for SIT here?} 
Overall, the scores suggest that more work is needed on multilingual adaptation of LLMs to unlock their ICL capabilities in other languages, and current adaptation strategies do not yield models with competitive (nor even useful at all) NLU. 
%that for ICL for multilingual NLU it is crucial that \textit{multilingual} data is included in pretraining as well as effective instruction tuning. 

\setlength{\tabcolsep}{4.5pt}
\begin{table}[!t]
\def\arraystretch{0.81}
{\fontsize{6.9}{7.1}\selectfont
%\footnotesize
%\resizebox{\linewidth}{!}{%
\begin{tabularx}{0.99\linewidth}{l XXXXX}
\toprule
\rowcolor{Gray} \textit{Model} & \textsc{am} & \textsc{en} & \textsc{mr} & \textsc{es} & \textsc{tr}                               \\ \cmidrule{2-6}
ICL: LLaMA-2  &   0.0  & 0.0 & 0.0  & 0.0 &  0.0          \\
ICL: MaLA-500   &   1.0  & 3.01 & 0.0  & 1.0 & 3.01                 \\ \cmidrule{2-6}
ICL: GPT-3.5   &  19.19 & 64.22 & 48.28 & 58.25 & 46.12                  \\
SIT: mT0   &  26.13 & 68.00 & 51.44 & 61.69 & 51.60                  \\ \cmidrule{2-6}
Tr-Test + ICL: GPT-3.5   &  32.25 & -- &  48.49 & 47.95 &   48.60                \\
 \bottomrule
\end{tabularx}
}
%}
\vspace{-1mm}
\caption{ICL results on the \textsc{id} task in the in-language in-domain setup.} \label{tab:icl_llama_vs_mala}
\vspace{-2.5mm}
\end{table}

\section{Conclusions and Future Work}
This work has provided a series of in-depth analyses of multilingual capabilities of three learning paradigms, two supervised ones versus in-context learning (ICL), with the focus on few-shot learning and NLU tasks. Besides task performance, the focus of the analyses has also been on multiple practical aspects (e.g., data efficiency, memory requirements, inference latency). As some of the key findings, we highlight that supervised approaches outperform ICL, even when substantially larger LLMs with higher inference costs are used for ICL. In addition, the analysis of target language adaptation on top of standard LLMs also does not paint a bright picture for multilingual NLP at the moment: while fluency of generated output improves post-adaptation, the output coherence and usefulness remains limited, plus the adapted LLMs lag substantially behind other (weakly supervised) approaches in NLU tasks for target languages.

%% (IV, return to CR)
%although language adaptation improves the model's ability to generate text in the target language, its understanding of the language only improves marginally. 

%% , and that ICL is currently not a viable solution for unseen NLU tasks for a wide range of languages. 

In general, our work has affirmed the importance of multilingual pretraining and the potential of supervised training on top of LLMs. Future work should invest more effort into the creation of massively multilingual- and multitask-pretrained LLMs with higher language coverage. Further, our analysis in \S\ref{sec:adaptation} calls for new and improved language adaptation methods atop the LLMs. 

%% (IV, return to CR)
%Finally, we hope that our work will also steer researchers and practitioners in multilingual NLP towards a more holistic view of model properties that combines task performance with practical aspects during the model development cycle (e.g., size, latency, memory, data collection cost). 

%demonstrates that future evaluation of the models and LLMs should cover not only tasks likely not seen in pretraining but also practical aspects such as inference time or cost so that more informed choice of a model could be used.

\section*{Acknowledgments}
The work has been in part supported by a Huawei research donation to the Language Technology Lab at
the University of Cambridge. It has also been supported by the UK Research and Innovation (UKRI) Frontier Research Grant EP/Y031350/1 EQUATE (the UK government’s funding guarantee for ERC Advanced Grants) awarded to Anna Korhonen at the University of Cambridge. The work of Ivan Vuli\'{c} has been supported in part by a Royal Society University Research Fellowship \textit{‘Inclusive and Sustainable Language Technology for a Truly Multilingual World’} (no 221137; 2022-).

%\nocite{Ando2005,borschinger-johnson-2011-particle,andrew2007scalable,rasooli-tetrault-2015,goodman-etal-2016-noise,harper-2014-learning}

\bibliography{anthology,custom}

\begin{thebibliography}{69}
\expandafter\ifx\csname natexlab\endcsname\relax\def\natexlab#1{#1}\fi

\bibitem[{Achiam et~al.(2023)Achiam, Adler, Agarwal, Ahmad, Akkaya, Aleman, Almeida, Altenschmidt, Altman, Anadkat et~al.}]{achiam2023gpt4technical}
Josh Achiam, Steven Adler, Sandhini Agarwal, Lama Ahmad, Ilge Akkaya, Florencia~Leoni Aleman, Diogo Almeida, Janko Altenschmidt, Sam Altman, Shyamal Anadkat, et~al. 2023.
\newblock \href {https://arxiv.org/abs/2303.08774} {Gpt-4 technical report}.
\newblock \emph{ArXiv preprint}, abs/2303.08774.

\bibitem[{Alizadeh et~al.(2023)Alizadeh, Mirzadeh, Belenko, Khatamifard, Cho, Del~Mundo, Rastegari, and Farajtabar}]{alizadeh2023llm}
Keivan Alizadeh, Iman Mirzadeh, Dmitry Belenko, Karen Khatamifard, Minsik Cho, Carlo~C Del~Mundo, Mohammad Rastegari, and Mehrdad Farajtabar. 2023.
\newblock \href {https://arxiv.org/abs/2312.11514} {Llm in a flash: Efficient large language model inference with limited memory}.
\newblock \emph{ArXiv preprint}, abs/2312.11514.

\bibitem[{Aminabadi et~al.(2022)Aminabadi, Rajbhandari, Awan, Li, Li, Zheng, Ruwase, Smith, Zhang, Rasley et~al.}]{aminabadi2022deepspeed}
Reza~Yazdani Aminabadi, Samyam Rajbhandari, Ammar~Ahmad Awan, Cheng Li, Du~Li, Elton Zheng, Olatunji Ruwase, Shaden Smith, Minjia Zhang, Jeff Rasley, et~al. 2022.
\newblock Deepspeed-inference: enabling efficient inference of transformer models at unprecedented scale.
\newblock In \emph{SC22: International Conference for High Performance Computing, Networking, Storage and Analysis}, pages 1--15. IEEE.

\bibitem[{Ansell et~al.(2021)Ansell, Ponti, Pfeiffer, Ruder, Glava{\v{s}}, Vuli{\'c}, and Korhonen}]{ansell-etal-2021-mad-g}
Alan Ansell, Edoardo~Maria Ponti, Jonas Pfeiffer, Sebastian Ruder, Goran Glava{\v{s}}, Ivan Vuli{\'c}, and Anna Korhonen. 2021.
\newblock \href {https://doi.org/10.18653/v1/2021.findings-emnlp.410} {{MAD}-{G}: {M}ultilingual adapter generation for efficient cross-lingual transfer}.
\newblock In \emph{Findings of the Association for Computational Linguistics: EMNLP 2021}, pages 4762--4781, Punta Cana, Dominican Republic. Association for Computational Linguistics.

\bibitem[{Balloccu et~al.(2024)Balloccu, Schmidtov{\'a}, Lango, and Du{\v{s}}ek}]{balloccu2024leak}
Simone Balloccu, Patr{\'\i}cia Schmidtov{\'a}, Mateusz Lango, and Ond{\v{r}}ej Du{\v{s}}ek. 2024.
\newblock \href {https://arxiv.org/abs/2402.03927} {Leak, cheat, repeat: Data contamination and evaluation malpractices in closed-source llms}.
\newblock \emph{ArXiv preprint}, abs/2402.03927.

\bibitem[{Bandarkar et~al.(2023)Bandarkar, Liang, Muller, Artetxe, Shukla, Husa, Goyal, Krishnan, Zettlemoyer, and Khabsa}]{Bandarkar:2023belebele}
Lucas Bandarkar, Davis Liang, Benjamin Muller, Mikel Artetxe, Satya~Narayan Shukla, Donald Husa, Naman Goyal, Abhinandan Krishnan, Luke Zettlemoyer, and Madian Khabsa. 2023.
\newblock \href {https://arxiv.org/abs/2308.16884} {{The Belebele Benchmark: a} parallel reading comprehension dataset in 122 language variants}.
\newblock \emph{ArXiv preprint}, abs/2308.16884.

\bibitem[{Bender(2011)}]{bender2011achieving}
Emily~M Bender. 2011.
\newblock On achieving and evaluating language-independence in nlp.
\newblock \emph{Linguistic Issues in Language Technology}, 6.

\bibitem[{Blevins and Zettlemoyer(2022)}]{blevins2022language-contamination}
Terra Blevins and Luke Zettlemoyer. 2022.
\newblock \href {https://arxiv.org/abs/2204.08110} {Language contamination helps explain the cross-lingual capabilities of english pretrained models}.
\newblock \emph{ArXiv preprint}, abs/2204.08110.

\bibitem[{Briakou et~al.(2023)Briakou, Cherry, and Foster}]{briakou-etal-2023-searching}
Eleftheria Briakou, Colin Cherry, and George Foster. 2023.
\newblock \href {https://doi.org/10.18653/v1/2023.acl-long.524} {Searching for needles in a haystack: On the role of incidental bilingualism in {P}a{LM}{'}s translation capability}.
\newblock In \emph{Proceedings of the 61st Annual Meeting of the Association for Computational Linguistics (Volume 1: Long Papers)}, pages 9432--9452, Toronto, Canada. Association for Computational Linguistics.

\bibitem[{Brown et~al.(2020)Brown, Mann, Ryder, Subbiah, Kaplan, Dhariwal, Neelakantan, Shyam, Sastry, Askell, Agarwal, Herbert{-}Voss, Krueger, Henighan, Child, Ramesh, Ziegler, Wu, Winter, Hesse, Chen, Sigler, Litwin, Gray, Chess, Clark, Berner, McCandlish, Radford, Sutskever, and Amodei}]{brown2020-in-context-learning}
Tom~B. Brown, Benjamin Mann, Nick Ryder, Melanie Subbiah, Jared Kaplan, Prafulla Dhariwal, Arvind Neelakantan, Pranav Shyam, Girish Sastry, Amanda Askell, Sandhini Agarwal, Ariel Herbert{-}Voss, Gretchen Krueger, Tom Henighan, Rewon Child, Aditya Ramesh, Daniel~M. Ziegler, Jeffrey Wu, Clemens Winter, Christopher Hesse, Mark Chen, Eric Sigler, Mateusz Litwin, Scott Gray, Benjamin Chess, Jack Clark, Christopher Berner, Sam McCandlish, Alec Radford, Ilya Sutskever, and Dario Amodei. 2020.
\newblock \href {https://proceedings.neurips.cc/paper/2020/hash/1457c0d6bfcb4967418bfb8ac142f64a-Abstract.html} {Language models are few-shot learners}.
\newblock In \emph{Advances in Neural Information Processing Systems 33: Annual Conference on Neural Information Processing Systems 2020, NeurIPS 2020, December 6-12, 2020, virtual}.

\bibitem[{Casanueva et~al.(2022)Casanueva, Vuli{\'c}, Spithourakis, and Budzianowski}]{casanueva-etal-2022-nluplusplus}
Inigo Casanueva, Ivan Vuli{\'c}, Georgios Spithourakis, and Pawe{\l} Budzianowski. 2022.
\newblock \href {https://doi.org/10.18653/v1/2022.findings-naacl.154} {{NLU}++: A multi-label, slot-rich, generalisable dataset for natural language understanding in task-oriented dialogue}.
\newblock In \emph{Findings of the Association for Computational Linguistics: NAACL 2022}, pages 1998--2013, Seattle, United States. Association for Computational Linguistics.

\bibitem[{Chen et~al.(2023)Chen, Chen, Zhu, and Zhou}]{chen-etal-2023-many}
Jiuhai Chen, Lichang Chen, Chen Zhu, and Tianyi Zhou. 2023.
\newblock \href {https://doi.org/10.18653/v1/2023.findings-emnlp.745} {How many demonstrations do you need for in-context learning?}
\newblock In \emph{Findings of the Association for Computational Linguistics: EMNLP 2023}, pages 11149--11159, Singapore. Association for Computational Linguistics.

\bibitem[{Chen et~al.(2022)Chen, Zhong, Zha, Karypis, and He}]{chen-etal-2022-meta}
Yanda Chen, Ruiqi Zhong, Sheng Zha, George Karypis, and He~He. 2022.
\newblock \href {https://doi.org/10.18653/v1/2022.acl-long.53} {Meta-learning via language model in-context tuning}.
\newblock In \emph{Proceedings of the 60th Annual Meeting of the Association for Computational Linguistics (Volume 1: Long Papers)}, pages 719--730, Dublin, Ireland. Association for Computational Linguistics.

\bibitem[{Chung et~al.(2022)Chung, Hou, Longpre, Zoph, Tay, Fedus, Li, Wang, Dehghani, Brahma et~al.}]{chung2022-flan}
Hyung~Won Chung, Le~Hou, Shayne Longpre, Barret Zoph, Yi~Tay, William Fedus, Yunxuan Li, Xuezhi Wang, Mostafa Dehghani, Siddhartha Brahma, et~al. 2022.
\newblock \href {https://arxiv.org/abs/2210.11416} {Scaling instruction-finetuned language models}.
\newblock \emph{ArXiv preprint}, abs/2210.11416.

\bibitem[{Conneau et~al.(2020)Conneau, Khandelwal, Goyal, Chaudhary, Wenzek, Guzm{\'a}n, Grave, Ott, Zettlemoyer, and Stoyanov}]{conneau-etal-2020-unsupervised}
Alexis Conneau, Kartikay Khandelwal, Naman Goyal, Vishrav Chaudhary, Guillaume Wenzek, Francisco Guzm{\'a}n, Edouard Grave, Myle Ott, Luke Zettlemoyer, and Veselin Stoyanov. 2020.
\newblock \href {https://doi.org/10.18653/v1/2020.acl-main.747} {Unsupervised cross-lingual representation learning at scale}.
\newblock In \emph{Proceedings of the 58th Annual Meeting of the Association for Computational Linguistics}, pages 8440--8451, Online. Association for Computational Linguistics.

\bibitem[{Conneau et~al.(2018)Conneau, Rinott, Lample, Williams, Bowman, Schwenk, and Stoyanov}]{conneau-etal-2018-xnli}
Alexis Conneau, Ruty Rinott, Guillaume Lample, Adina Williams, Samuel Bowman, Holger Schwenk, and Veselin Stoyanov. 2018.
\newblock \href {https://doi.org/10.18653/v1/D18-1269} {{XNLI}: Evaluating cross-lingual sentence representations}.
\newblock In \emph{Proceedings of the 2018 Conference on Empirical Methods in Natural Language Processing}, pages 2475--2485, Brussels, Belgium. Association for Computational Linguistics.

\bibitem[{Cui et~al.(2023)Cui, Yang, and Yao}]{cui2023efficient}
Yiming Cui, Ziqing Yang, and Xin Yao. 2023.
\newblock \href {https://arxiv.org/abs/2304.08177} {Efficient and effective text encoding for chinese llama and alpaca}.
\newblock \emph{ArXiv preprint}, abs/2304.08177.

\bibitem[{Dettmers et~al.(2023)Dettmers, Pagnoni, Holtzman, and Zettlemoyer}]{dettmers2023qlora}
Tim Dettmers, Artidoro Pagnoni, Ari Holtzman, and Luke Zettlemoyer. 2023.
\newblock \href {https://arxiv.org/abs/2305.14314} {Qlora: Efficient finetuning of quantized llms}.
\newblock \emph{ArXiv preprint}, abs/2305.14314.

\bibitem[{Devlin et~al.(2019)Devlin, Chang, Lee, and Toutanova}]{devlin-etal-2019-bert}
Jacob Devlin, Ming-Wei Chang, Kenton Lee, and Kristina Toutanova. 2019.
\newblock \href {https://doi.org/10.18653/v1/N19-1423} {{BERT}: Pre-training of deep bidirectional transformers for language understanding}.
\newblock In \emph{Proceedings of the 2019 Conference of the North {A}merican Chapter of the Association for Computational Linguistics: Human Language Technologies, Volume 1 (Long and Short Papers)}, pages 4171--4186, Minneapolis, Minnesota. Association for Computational Linguistics.

\bibitem[{Doddapaneni et~al.(2021)Doddapaneni, Ramesh, Khapra, Kunchukuttan, and Kumar}]{doddapaneni2021primer}
Sumanth Doddapaneni, Gowtham Ramesh, Mitesh~M Khapra, Anoop Kunchukuttan, and Pratyush Kumar. 2021.
\newblock \href {https://arxiv.org/abs/2107.00676} {A primer on pretrained multilingual language models}.
\newblock \emph{ArXiv preprint}, abs/2107.00676.

\bibitem[{Feng et~al.(2022)Feng, Yang, Cer, Arivazhagan, and Wang}]{feng-etal-2022-labse}
Fangxiaoyu Feng, Yinfei Yang, Daniel Cer, Naveen Arivazhagan, and Wei Wang. 2022.
\newblock \href {https://doi.org/10.18653/v1/2022.acl-long.62} {Language-agnostic {BERT} sentence embedding}.
\newblock In \emph{Proceedings of the 60th Annual Meeting of the Association for Computational Linguistics (Volume 1: Long Papers)}, pages 878--891, Dublin, Ireland. Association for Computational Linguistics.

\bibitem[{Fuisz et~al.(2022)Fuisz, Vulic, Gibbons, Casanueva, and Budzianowski}]{Fuisz2022improvedve-qa}
Gabor Fuisz, Ivan Vulic, Samuel Gibbons, I{\~n}igo Casanueva, and Paweł Budzianowski. 2022.
\newblock \href {https://arxiv.org/abs/2204.02123} {Improved and efficient conversational slot labeling through question answering}.
\newblock \emph{ArXiv preprint}, abs/2204.02123.

\bibitem[{Han et~al.(2023)Han, Simig, Mihaylov, Tsvetkov, Celikyilmaz, and Wang}]{han-etal-2023-understanding}
Xiaochuang Han, Daniel Simig, Todor Mihaylov, Yulia Tsvetkov, Asli Celikyilmaz, and Tianlu Wang. 2023.
\newblock \href {https://doi.org/10.18653/v1/2023.acl-long.708} {Understanding in-context learning via supportive pretraining data}.
\newblock In \emph{Proceedings of the 61st Annual Meeting of the Association for Computational Linguistics (Volume 1: Long Papers)}, pages 12660--12673, Toronto, Canada. Association for Computational Linguistics.

\bibitem[{Hu et~al.(2022)Hu, Shen, Wallis, Allen{-}Zhu, Li, Wang, Wang, and Chen}]{hu2022lora}
Edward~J. Hu, Yelong Shen, Phillip Wallis, Zeyuan Allen{-}Zhu, Yuanzhi Li, Shean Wang, Lu~Wang, and Weizhu Chen. 2022.
\newblock \href {https://openreview.net/forum?id=nZeVKeeFYf9} {Lora: Low-rank adaptation of large language models}.
\newblock In \emph{The Tenth International Conference on Learning Representations, {ICLR} 2022, Virtual Event, April 25-29, 2022}. OpenReview.net.

\bibitem[{Hu et~al.(2023{\natexlab{a}})Hu, Zhou, Hergul, Gritta, Zhang, Iacobacci, Vuli{\'c}, and Korhonen}]{hu2023multi3woz}
Songbo Hu, Han Zhou, Mete Hergul, Milan Gritta, Guchun Zhang, Ignacio Iacobacci, Ivan Vuli{\'c}, and Anna Korhonen. 2023{\natexlab{a}}.
\newblock Multi 3 woz: A multilingual, multi-domain, multi-parallel dataset for training and evaluating culturally adapted task-oriented dialog systems.
\newblock \emph{Transactions of the Association for Computational Linguistics}, 11:1396--1415.

\bibitem[{Hu et~al.(2023{\natexlab{b}})Hu, Zhou, Yuan, Gritta, Zhang, Iacobacci, Korhonen, and Vuli{\'c}}]{hu-etal-2023-systematic}
Songbo Hu, Han Zhou, Moy Yuan, Milan Gritta, Guchun Zhang, Ignacio Iacobacci, Anna Korhonen, and Ivan Vuli{\'c}. 2023{\natexlab{b}}.
\newblock \href {https://doi.org/10.18653/v1/2023.emnlp-main.422} {A systematic study of performance disparities in multilingual task-oriented dialogue systems}.
\newblock In \emph{Proceedings of the 2023 Conference on Empirical Methods in Natural Language Processing}, pages 6825--6851, Singapore. Association for Computational Linguistics.

\bibitem[{Huyen(2022)}]{huyen2022designingmlmsys-latency}
Chip Huyen. 2022.
\newblock \emph{Designing machine learning systems}.
\newblock " O'Reilly Media, Inc.".

\bibitem[{ImaniGooghari et~al.(2023)ImaniGooghari, Lin, Kargaran, Severini, Jalili~Sabet, Kassner, Ma, Schmid, Martins, Yvon, and Sch{\"u}tze}]{imanigooghari-etal-2023-glot500}
Ayyoob ImaniGooghari, Peiqin Lin, Amir~Hossein Kargaran, Silvia Severini, Masoud Jalili~Sabet, Nora Kassner, Chunlan Ma, Helmut Schmid, Andr{\'e} Martins, Fran{\c{c}}ois Yvon, and Hinrich Sch{\"u}tze. 2023.
\newblock \href {https://doi.org/10.18653/v1/2023.acl-long.61} {Glot500: Scaling multilingual corpora and language models to 500 languages}.
\newblock In \emph{Proceedings of the 61st Annual Meeting of the Association for Computational Linguistics (Volume 1: Long Papers)}, pages 1082--1117, Toronto, Canada. Association for Computational Linguistics.

\bibitem[{Kargaran et~al.(2023)Kargaran, Imani, Yvon, and Schuetze}]{kargaran-etal-2023-glotlid}
Amir Kargaran, Ayyoob Imani, Fran{\c{c}}ois Yvon, and Hinrich Schuetze. 2023.
\newblock \href {https://doi.org/10.18653/v1/2023.findings-emnlp.410} {{G}lot{LID}: Language identification for low-resource languages}.
\newblock In \emph{Findings of the Association for Computational Linguistics: EMNLP 2023}, pages 6155--6218, Singapore. Association for Computational Linguistics.

\bibitem[{Kew et~al.(2023)Kew, Schottmann, and Sennrich}]{kew2023turning}
Tannon Kew, Florian Schottmann, and Rico Sennrich. 2023.
\newblock \href {https://arxiv.org/abs/2312.12683} {Turning english-centric llms into polyglots: How much multilinguality is needed?}
\newblock \emph{ArXiv preprint}, abs/2312.12683.

\bibitem[{Lauscher et~al.(2020)Lauscher, Ravishankar, Vuli{\'c}, and Glava{\v{s}}}]{lauscher2020zero-to-hero}
Anne Lauscher, Vinit Ravishankar, Ivan Vuli{\'c}, and Goran Glava{\v{s}}. 2020.
\newblock \href {https://doi.org/10.18653/v1/2020.emnlp-main.363} {From zero to hero: {O}n the limitations of zero-shot language transfer with multilingual {T}ransformers}.
\newblock In \emph{Proceedings of the 2020 Conference on Empirical Methods in Natural Language Processing (EMNLP)}, pages 4483--4499, Online. Association for Computational Linguistics.

\bibitem[{Li et~al.(2023{\natexlab{a}})Li, Zhou, Glavaš, Korhonen, and Vulić}]{chengzu2023}
Chengzu Li, Han Zhou, Goran Glavaš, Anna Korhonen, and Ivan Vulić. 2023{\natexlab{a}}.
\newblock \href {http://arxiv.org/abs/2312.13772} {On task performance and model calibration with supervised and self-ensembled in-context learning}.

\bibitem[{Li et~al.(2023{\natexlab{b}})Li, Koto, Wu, Aji, and Baldwin}]{li2023bactrian}
Haonan Li, Fajri Koto, Minghao Wu, Alham~Fikri Aji, and Timothy Baldwin. 2023{\natexlab{b}}.
\newblock \href {https://arxiv.org/abs/2305.15011} {Bactrian-x: A multilingual replicable instruction-following model with low-rank adaptation}.
\newblock \emph{ArXiv preprint}, abs/2305.15011.

\bibitem[{Lin(2004)}]{lin-2004-rouge}
Chin-Yew Lin. 2004.
\newblock \href {https://aclanthology.org/W04-1013} {{ROUGE}: A package for automatic evaluation of summaries}.
\newblock In \emph{Text Summarization Branches Out}, pages 74--81, Barcelona, Spain. Association for Computational Linguistics.

\bibitem[{Lin et~al.(2024)Lin, Ji, Tiedemann, Martins, and Sch{\"u}tze}]{lin2024mala500}
Peiqin Lin, Shaoxiong Ji, J{\"o}rg Tiedemann, Andr{\'e}~FT Martins, and Hinrich Sch{\"u}tze. 2024.
\newblock \href {https://arxiv.org/abs/2401.13303} {Mala-500: Massive language adaptation of large language models}.
\newblock \emph{ArXiv preprint}, abs/2401.13303.

\bibitem[{Lin et~al.(2022)Lin, Mihaylov, Artetxe, Wang, Chen, Simig, Ott, Goyal, Bhosale, Du, Pasunuru, Shleifer, Koura, Chaudhary, O{'}Horo, Wang, Zettlemoyer, Kozareva, Diab, Stoyanov, and Li}]{lin-etal-2022-shot}
Xi~Victoria Lin, Todor Mihaylov, Mikel Artetxe, Tianlu Wang, Shuohui Chen, Daniel Simig, Myle Ott, Naman Goyal, Shruti Bhosale, Jingfei Du, Ramakanth Pasunuru, Sam Shleifer, Punit~Singh Koura, Vishrav Chaudhary, Brian O{'}Horo, Jeff Wang, Luke Zettlemoyer, Zornitsa Kozareva, Mona Diab, Veselin Stoyanov, and Xian Li. 2022.
\newblock \href {https://aclanthology.org/2022.emnlp-main.616} {Few-shot learning with multilingual generative language models}.
\newblock In \emph{Proceedings of the 2022 Conference on Empirical Methods in Natural Language Processing}, pages 9019--9052, Abu Dhabi, United Arab Emirates. Association for Computational Linguistics.

\bibitem[{Liu et~al.(2023)Liu, Lin, Hewitt, Paranjape, Bevilacqua, Petroni, and Liang}]{lostinthemiddle}
Nelson~F. Liu, Kevin Lin, John Hewitt, Ashwin Paranjape, Michele Bevilacqua, Fabio Petroni, and Percy Liang. 2023.
\newblock \href {https://arxiv.org/abs/2307.03172} {Lost in the middle: How language models use long contexts}.
\newblock \emph{ArXiv preprint}, abs/2307.03172.

\bibitem[{Loshchilov and Hutter(2019)}]{adamw}
Ilya Loshchilov and Frank Hutter. 2019.
\newblock \href {https://openreview.net/forum?id=Bkg6RiCqY7} {Decoupled weight decay regularization}.
\newblock In \emph{7th International Conference on Learning Representations, {ICLR} 2019, New Orleans, LA, USA, May 6-9, 2019}. OpenReview.net.

\bibitem[{Min et~al.(2022)Min, Lewis, Zettlemoyer, and Hajishirzi}]{min-etal-2022-metaicl}
Sewon Min, Mike Lewis, Luke Zettlemoyer, and Hannaneh Hajishirzi. 2022.
\newblock \href {https://doi.org/10.18653/v1/2022.naacl-main.201} {{M}eta{ICL}: Learning to learn in context}.
\newblock In \emph{Proceedings of the 2022 Conference of the North American Chapter of the Association for Computational Linguistics: Human Language Technologies}, pages 2791--2809, Seattle, United States. Association for Computational Linguistics.

\bibitem[{Mishra et~al.(2022{\natexlab{a}})Mishra, Khashabi, Baral, and Hajishirzi}]{mishra2022cross}
Swaroop Mishra, Daniel Khashabi, Chitta Baral, and Hannaneh Hajishirzi. 2022{\natexlab{a}}.
\newblock \href {https://doi.org/10.18653/v1/2022.acl-long.244} {Cross-task generalization via natural language crowdsourcing instructions}.
\newblock In \emph{Proceedings of the 60th Annual Meeting of the Association for Computational Linguistics (Volume 1: Long Papers)}, pages 3470--3487, Dublin, Ireland. Association for Computational Linguistics.

\bibitem[{Mishra et~al.(2022{\natexlab{b}})Mishra, Khashabi, Baral, and Hajishirzi}]{mishra-etal-2022-cross}
Swaroop Mishra, Daniel Khashabi, Chitta Baral, and Hannaneh Hajishirzi. 2022{\natexlab{b}}.
\newblock \href {https://doi.org/10.18653/v1/2022.acl-long.244} {Cross-task generalization via natural language crowdsourcing instructions}.
\newblock In \emph{Proceedings of the 60th Annual Meeting of the Association for Computational Linguistics (Volume 1: Long Papers)}, pages 3470--3487, Dublin, Ireland. Association for Computational Linguistics.

\bibitem[{Moghe et~al.(2023)Moghe, Razumovskaia, Guillou, Vuli{\'c}, Korhonen, and Birch}]{moghe-etal-2023-multi3nlu}
Nikita Moghe, Evgeniia Razumovskaia, Liane Guillou, Ivan Vuli{\'c}, Anna Korhonen, and Alexandra Birch. 2023.
\newblock \href {https://doi.org/10.18653/v1/2023.findings-acl.230} {{M}ulti3{NLU}++: A multilingual, multi-intent, multi-domain dataset for natural language understanding in task-oriented dialogue}.
\newblock In \emph{Findings of the Association for Computational Linguistics: ACL 2023}, pages 3732--3755, Toronto, Canada. Association for Computational Linguistics.

\bibitem[{Muennighoff et~al.(2022)Muennighoff, Wang, Sutawika, Roberts, Biderman, Scao, Bari, Shen, Yong, Schoelkopf et~al.}]{muennighoff2022-mt0}
Niklas Muennighoff, Thomas Wang, Lintang Sutawika, Adam Roberts, Stella Biderman, Teven~Le Scao, M~Saiful Bari, Sheng Shen, Zheng-Xin Yong, Hailey Schoelkopf, et~al. 2022.
\newblock \href {https://arxiv.org/abs/2211.01786} {Crosslingual generalization through multitask finetuning}.
\newblock \emph{ArXiv preprint}, abs/2211.01786.

\bibitem[{Ojo et~al.(2023)Ojo, Ogueji, Stenetorp, and Adelani}]{ojo2023good}
Jessica Ojo, Kelechi Ogueji, Pontus Stenetorp, and David~I Adelani. 2023.
\newblock \href {https://arxiv.org/abs/2311.07978} {How good are large language models on african languages?}
\newblock \emph{ArXiv preprint}, abs/2311.07978.

\bibitem[{Page-Caccia et~al.(2024)Page-Caccia, Ponti, Su, Pereira, Le~Roux, and Sordoni}]{page2024crosstask-peft}
Lucas Page-Caccia, Edoardo~Maria Ponti, Zhan Su, Matheus Pereira, Nicolas Le~Roux, and Alessandro Sordoni. 2024.
\newblock Multi-head adapter routing for cross-task generalization.
\newblock \emph{Advances in Neural Information Processing Systems}, 36.

\bibitem[{Papineni et~al.(2002)Papineni, Roukos, Ward, and Zhu}]{Papineni-2002-bleu}
Kishore Papineni, Salim Roukos, Todd Ward, and Wei-Jing Zhu. 2002.
\newblock \href {https://doi.org/10.3115/1073083.1073135} {{B}leu: a method for automatic evaluation of machine translation}.
\newblock In \emph{Proceedings of the 40th Annual Meeting of the Association for Computational Linguistics}, pages 311--318, Philadelphia, Pennsylvania, USA. Association for Computational Linguistics.

\bibitem[{Pillutla et~al.(2021)Pillutla, Swayamdipta, Zellers, Thickstun, Welleck, Choi, and Harchaoui}]{pillutla-etal:mauve:neurips2021}
Krishna Pillutla, Swabha Swayamdipta, Rowan Zellers, John Thickstun, Sean Welleck, Yejin Choi, and Za{\"{\i}}d Harchaoui. 2021.
\newblock \href {https://proceedings.neurips.cc/paper/2021/hash/260c2432a0eecc28ce03c10dadc078a4-Abstract.html} {{MAUVE:} measuring the gap between neural text and human text using divergence frontiers}.
\newblock In \emph{Advances in Neural Information Processing Systems 34: Annual Conference on Neural Information Processing Systems 2021, NeurIPS 2021, December 6-14, 2021, virtual}, pages 4816--4828.

\bibitem[{Press et~al.(2022)Press, Smith, and Lewis}]{alibi}
Ofir Press, Noah~A. Smith, and Mike Lewis. 2022.
\newblock \href {https://openreview.net/forum?id=R8sQPpGCv0} {Train short, test long: Attention with linear biases enables input length extrapolation}.
\newblock In \emph{The Tenth International Conference on Learning Representations, {ICLR} 2022, Virtual Event, April 25-29, 2022}. OpenReview.net.

\bibitem[{Radford et~al.(2019)Radford, Wu, Child, Luan, Amodei, Sutskever et~al.}]{radford2019-in-context-learning}
Alec Radford, Jeffrey Wu, Rewon Child, David Luan, Dario Amodei, Ilya Sutskever, et~al. 2019.
\newblock Language models are unsupervised multitask learners.
\newblock \emph{OpenAI blog}, 1(8):9.

\bibitem[{Razumovskaia et~al.(2023)Razumovskaia, Glava{\v{s}}, Korhonen, and Vuli{\'c}}]{razumovskaia2023sqatin}
Evgeniia Razumovskaia, Goran Glava{\v{s}}, Anna Korhonen, and Ivan Vuli{\'c}. 2023.
\newblock \href {https://arxiv.org/abs/2311.09502} {Sqatin: Supervised instruction tuning meets question answering for improved dialogue nlu}.
\newblock \emph{ArXiv preprint}, abs/2311.09502.

\bibitem[{Rubin and Berant(2023)}]{Rubin:2023arxiv}
Ohad Rubin and Jonathan Berant. 2023.
\newblock \href {https://arxiv.org/abs/2306.13421} {Long-range language modeling with self-retrieval}.
\newblock \emph{ArXiv preprint}, abs/2306.13421.

\bibitem[{Ruder et~al.(2023)Ruder, Clark, Gutkin, Kale, Ma, Nicosia, Rijhwani, Riley, Sarr, Wang, Wieting, Gupta, Katanova, Kirov, Dickinson, Roark, Samanta, Tao, Adelani, Axelrod, Caswell, Cherry, Garrette, Ingle, Johnson, Panteleev, and Talukdar}]{ruder-etal-2023-xtremeup}
Sebastian Ruder, Jonathan Clark, Alexander Gutkin, Mihir Kale, Min Ma, Massimo Nicosia, Shruti Rijhwani, Parker Riley, Jean-Michel Sarr, Xinyi Wang, John Wieting, Nitish Gupta, Anna Katanova, Christo Kirov, Dana Dickinson, Brian Roark, Bidisha Samanta, Connie Tao, David Adelani, Vera Axelrod, Isaac Caswell, Colin Cherry, Dan Garrette, Reeve Ingle, Melvin Johnson, Dmitry Panteleev, and Partha Talukdar. 2023.
\newblock \href {https://doi.org/10.18653/v1/2023.findings-emnlp.125} {{XTREME}-{UP}: A user-centric scarce-data benchmark for under-represented languages}.
\newblock In \emph{Findings of the Association for Computational Linguistics: EMNLP 2023}, pages 1856--1884, Singapore. Association for Computational Linguistics.

\bibitem[{Sainz et~al.(2023)Sainz, Campos, Garc{\'\i}a-Ferrero, Etxaniz, de~Lacalle, and Agirre}]{sainz2023nlp}
Oscar Sainz, Jon Campos, Iker Garc{\'\i}a-Ferrero, Julen Etxaniz, Oier~Lopez de~Lacalle, and Eneko Agirre. 2023.
\newblock Nlp evaluation in trouble: On the need to measure llm data contamination for each benchmark.
\newblock In \emph{Findings of the Association for Computational Linguistics: EMNLP 2023}, pages 10776--10787.

\bibitem[{Sanh et~al.(2022)Sanh, Webson, Raffel, Bach, Sutawika, Alyafeai, Chaffin, Stiegler, Raja, Dey, Bari, Xu, Thakker, Sharma, Szczechla, Kim, Chhablani, Nayak, Datta, Chang, Jiang, Wang, Manica, Shen, Yong, Pandey, Bawden, Wang, Neeraj, Rozen, Sharma, Santilli, F{\'{e}}vry, Fries, Teehan, Scao, Biderman, Gao, Wolf, and Rush}]{sanh2022multitask-t0}
Victor Sanh, Albert Webson, Colin Raffel, Stephen~H. Bach, Lintang Sutawika, Zaid Alyafeai, Antoine Chaffin, Arnaud Stiegler, Arun Raja, Manan Dey, M~Saiful Bari, Canwen Xu, Urmish Thakker, Shanya~Sharma Sharma, Eliza Szczechla, Taewoon Kim, Gunjan Chhablani, Nihal~V. Nayak, Debajyoti Datta, Jonathan Chang, Mike~Tian{-}Jian Jiang, Han Wang, Matteo Manica, Sheng Shen, Zheng~Xin Yong, Harshit Pandey, Rachel Bawden, Thomas Wang, Trishala Neeraj, Jos Rozen, Abheesht Sharma, Andrea Santilli, Thibault F{\'{e}}vry, Jason~Alan Fries, Ryan Teehan, Teven~Le Scao, Stella Biderman, Leo Gao, Thomas Wolf, and Alexander~M. Rush. 2022.
\newblock \href {https://openreview.net/forum?id=9Vrb9D0WI4} {Multitask prompted training enables zero-shot task generalization}.
\newblock In \emph{The Tenth International Conference on Learning Representations, {ICLR} 2022, Virtual Event, April 25-29, 2022}. OpenReview.net.

\bibitem[{Shaham et~al.(2024)Shaham, Herzig, Aharoni, Szpektor, Tsarfaty, and Eyal}]{shaham2024multilingual}
Uri Shaham, Jonathan Herzig, Roee Aharoni, Idan Szpektor, Reut Tsarfaty, and Matan Eyal. 2024.
\newblock \href {https://arxiv.org/abs/2401.01854} {Multilingual instruction tuning with just a pinch of multilinguality}.
\newblock \emph{ArXiv preprint}, abs/2401.01854.

\bibitem[{Shi et~al.(2022)Shi, Suzgun, Freitag, Wang, Srivats, Vosoughi, Chung, Tay, Ruder, Zhou et~al.}]{shi2022language-multilingual-cot}
Freda Shi, Mirac Suzgun, Markus Freitag, Xuezhi Wang, Suraj Srivats, Soroush Vosoughi, Hyung~Won Chung, Yi~Tay, Sebastian Ruder, Denny Zhou, et~al. 2022.
\newblock Language models are multilingual chain-of-thought reasoners.
\newblock In \emph{The Eleventh International Conference on Learning Representations}.

\bibitem[{Shliazhko et~al.(2022)Shliazhko, Fenogenova, Tikhonova, Mikhailov, Kozlova, and Shavrina}]{shliazhko2022mgpt}
Oleh Shliazhko, Alena Fenogenova, Maria Tikhonova, Vladislav Mikhailov, Anastasia Kozlova, and Tatiana Shavrina. 2022.
\newblock \href {https://arxiv.org/abs/2204.07580} {mgpt: Few-shot learners go multilingual}.
\newblock \emph{ArXiv preprint}, abs/2204.07580.

\bibitem[{Sitaram et~al.(2023)Sitaram, Choudhury, Patra, Chaudhary, Ahuja, and Bali}]{sitaram2023everything}
Sunayana Sitaram, Monojit Choudhury, Barun Patra, Vishrav Chaudhary, Kabir Ahuja, and Kalika Bali. 2023.
\newblock Everything you need to know about multilingual llms: Towards fair, performant and reliable models for languages of the world.
\newblock In \emph{Proceedings of the 61st Annual Meeting of the Association for Computational Linguistics (Volume 6: Tutorial Abstracts)}, pages 21--26.

\bibitem[{Tanwar et~al.(2023)Tanwar, Dutta, Borthakur, and Chakraborty}]{tanwar-etal-2023-multilingual}
Eshaan Tanwar, Subhabrata Dutta, Manish Borthakur, and Tanmoy Chakraborty. 2023.
\newblock \href {https://doi.org/10.18653/v1/2023.acl-long.346} {Multilingual {LLM}s are better cross-lingual in-context learners with alignment}.
\newblock In \emph{Proceedings of the 61st Annual Meeting of the Association for Computational Linguistics (Volume 1: Long Papers)}, pages 6292--6307, Toronto, Canada. Association for Computational Linguistics.

\bibitem[{Touvron et~al.(2023)Touvron, Martin, Stone, Albert, Almahairi, Babaei, Bashlykov, Batra, Bhargava, Bhosale et~al.}]{touvron2023llama2}
Hugo Touvron, Louis Martin, Kevin Stone, Peter Albert, Amjad Almahairi, Yasmine Babaei, Nikolay Bashlykov, Soumya Batra, Prajjwal Bhargava, Shruti Bhosale, et~al. 2023.
\newblock \href {https://arxiv.org/abs/2307.09288} {Llama 2: Open foundation and fine-tuned chat models}.
\newblock \emph{ArXiv preprint}, abs/2307.09288.

\bibitem[{Wei et~al.(2022{\natexlab{a}})Wei, Bosma, Zhao, Guu, Yu, Lester, Du, Dai, and Le}]{wei2021finetuned}
Jason Wei, Maarten Bosma, Vincent~Y. Zhao, Kelvin Guu, Adams~Wei Yu, Brian Lester, Nan Du, Andrew~M. Dai, and Quoc~V. Le. 2022{\natexlab{a}}.
\newblock \href {https://openreview.net/forum?id=gEZrGCozdqR} {Finetuned language models are zero-shot learners}.
\newblock In \emph{The Tenth International Conference on Learning Representations, {ICLR} 2022, Virtual Event, April 25-29, 2022}. OpenReview.net.

\bibitem[{Wei et~al.(2022{\natexlab{b}})Wei, Bosma, Zhao, Guu, Yu, Lester, Du, Dai, and Le}]{wei2022finetuned-sit-iclr}
Jason Wei, Maarten Bosma, Vincent~Y. Zhao, Kelvin Guu, Adams~Wei Yu, Brian Lester, Nan Du, Andrew~M. Dai, and Quoc~V. Le. 2022{\natexlab{b}}.
\newblock \href {https://openreview.net/forum?id=gEZrGCozdqR} {Finetuned language models are zero-shot learners}.
\newblock In \emph{The Tenth International Conference on Learning Representations, {ICLR} 2022, Virtual Event, April 25-29, 2022}. OpenReview.net.

\bibitem[{Wei et~al.(2022{\natexlab{c}})Wei, Tay, Bommasani, Raffel, Zoph, Borgeaud, Yogatama, Bosma, Zhou, Metzler, Chi, Hashimoto, Vinyals, Liang, Dean, and Fedus}]{Wei:2022emergent}
Jason Wei, Yi~Tay, Rishi Bommasani, Colin Raffel, Barret Zoph, Sebastian Borgeaud, Dani Yogatama, Maarten Bosma, Denny Zhou, Donald Metzler, Ed~H. Chi, Tatsunori Hashimoto, Oriol Vinyals, Percy Liang, Jeff Dean, and William Fedus. 2022{\natexlab{c}}.
\newblock \href {https://openreview.net/forum?id=yzkSU5zdwD} {Emergent abilities of large language models}.
\newblock \emph{Transactions on Machine Learning Research}, 2022.

\bibitem[{Wei et~al.(2023)Wei, Wei, Lin, Li, Zhang, Ren, Li, Wan, Cao, Xie et~al.}]{wei2023polylm}
Xiangpeng Wei, Haoran Wei, Huan Lin, Tianhao Li, Pei Zhang, Xingzhang Ren, Mei Li, Yu~Wan, Zhiwei Cao, Binbin Xie, et~al. 2023.
\newblock \href {https://arxiv.org/abs/2307.06018} {Polylm: An open source polyglot large language model}.
\newblock \emph{ArXiv preprint}, abs/2307.06018.

\bibitem[{Williams et~al.(2018)Williams, Nangia, and Bowman}]{williams-etal-2018-broad}
Adina Williams, Nikita Nangia, and Samuel Bowman. 2018.
\newblock \href {https://doi.org/10.18653/v1/N18-1101} {A broad-coverage challenge corpus for sentence understanding through inference}.
\newblock In \emph{Proceedings of the 2018 Conference of the North {A}merican Chapter of the Association for Computational Linguistics: Human Language Technologies, Volume 1 (Long Papers)}, pages 1112--1122, New Orleans, Louisiana. Association for Computational Linguistics.

\bibitem[{Winata et~al.(2021)Winata, Madotto, Lin, Liu, Yosinski, and Fung}]{winata2021language}
Genta~Indra Winata, Andrea Madotto, Zhaojiang Lin, Rosanne Liu, Jason Yosinski, and Pascale Fung. 2021.
\newblock \href {https://doi.org/10.18653/v1/2021.mrl-1.1} {Language models are few-shot multilingual learners}.
\newblock In \emph{Proceedings of the 1st Workshop on Multilingual Representation Learning}, pages 1--15, Punta Cana, Dominican Republic. Association for Computational Linguistics.

\bibitem[{Xue et~al.(2021)Xue, Constant, Roberts, Kale, Al-Rfou, Siddhant, Barua, and Raffel}]{xue-etal-2021-mt5}
Linting Xue, Noah Constant, Adam Roberts, Mihir Kale, Rami Al-Rfou, Aditya Siddhant, Aditya Barua, and Colin Raffel. 2021.
\newblock \href {https://doi.org/10.18653/v1/2021.naacl-main.41} {m{T}5: A massively multilingual pre-trained text-to-text transformer}.
\newblock In \emph{Proceedings of the 2021 Conference of the North American Chapter of the Association for Computational Linguistics: Human Language Technologies}, pages 483--498, Online. Association for Computational Linguistics.

\bibitem[{Ye et~al.(2023)Ye, Chen, Xu, Zu, Shao, Liu, Cui, Zhou, Gong, Shen et~al.}]{ye2023comprehensive}
Junjie Ye, Xuanting Chen, Nuo Xu, Can Zu, Zekai Shao, Shichun Liu, Yuhan Cui, Zeyang Zhou, Chao Gong, Yang Shen, et~al. 2023.
\newblock \href {https://arxiv.org/abs/2303.10420} {A comprehensive capability analysis of gpt-3 and gpt-3.5 series models}.
\newblock \emph{ArXiv preprint}, abs/2303.10420.

\bibitem[{Zhao et~al.(2024)Zhao, Zhang, Zhang, Gui, and Huang}]{zhao2024llama}
Jun Zhao, Zhihao Zhang, Qi~Zhang, Tao Gui, and Xuanjing Huang. 2024.
\newblock \href {https://arxiv.org/abs/2401.01055} {Llama beyond english: An empirical study on language capability transfer}.
\newblock \emph{ArXiv preprint}, abs/2401.01055.

\end{thebibliography}
\bibliographystyle{acl_natbib}

\iffalse
\section*{Limitations}

 * need to verbalise intent and slot classes for ICL 

 * medium-sized LLMs

\fi

\appendix
\clearpage

\iffalse
\section{Details of Evaluation Datasets}\label{app:dataset_stats} The summary of evaluation datasets is shown in Table \ref{tab:dataset_stats}.

\noindent Multi3NLU++ \cite{moghe-etal-2023-multi3nlu} is a multi-domain multili-intent multilingual dataset created via professional translation of English NLU++ \cite{casanueva-etal-2022-nluplusplus} dataset into the following 4 languages: Spanish (\textsc{es}), Turkish (\textsc{tr}), Marathi (\textsc{mr}) and Amharic (\textsc{am}). The languages are ordered from high- to low-resource. They belong to 3 different language families and use 3 different scripts. This diversity allows us to scrutinise the model's performance along different axes of linguistic diversity. 

\noindent XNLI \cite{conneau-etal-2018-xnli} is a multilingual natural language inference dataset where each example consists of hypothesis and premise. The model needs to determine the relationship between two sentences as either contradiction, entailmeant ($premise \rightarrow hypothesis$) or neutral with respect to the premise. The dataset covers 15 high- to medium-resource languages. We focus on 3 non-English languages, Spanish (\textsc{es}), Russian (\textsc{ru}) and Turkish (\textsc{tr}). As XNLI is a widely-used benchmark, it is likely that LLMs have seen the task in pretraining (potentially including some parts of the test set).
\fi

\section{Instructions Used for Different Tasks}
\label{sec:instructions_for_tasks}

\begin{table}[!th]
\footnotesize
\begin{tabularx}{0.45\textwidth}{lX}
\toprule
\textbf{Task} & \textbf{Instruction Text}           \\ \midrule
ID   & The aim is to understand user's intent from the utterance.\\   & Include all applicable options exactly as they are provided. Separate the classes \\   & by hyphen. \\    & If no options are applicable, return an empty string. \\   & Options:\\    & - to deny something\\    & - to ask about savings account \\   & \textless{}list of all options applicable in the domain\textgreater\\   &  \\    & Utterance: \{demonstration1\}\\    & Intents: \{intent1\}-\{intent2\}-\{intent3\}\\    & \textless{}all in-context demonstrations\textgreater\\    & \\   &  Utterance: \{test example\}\\    & Intents: \\ \hline
VE   & The aim is to extract slot values from the user utterance. \\    & Use \$\$ as delimiter between slot-value pairs.\\    & The slot values should be tagged as:\\    &  - amount\_of\_money: specific amount of money\\    &  - adults: number of adults\\    & \textless{}list of all slot classes applicable in a given domain\textgreater\\    & \\    & Utterance: \{demonstration1\} \\    & Values: \{slot\_class1\}:\{value1\}\$\$\{slot\_class2\}:\{value2\}\\    & \textless{}all in-context demonstrations\textgreater\\    & \\    & Utterance: \{test example\}\\    & Values:                                                                                               \\ \hline
NLI  & The aim is to determine whether the premise entails, contradicts or is neutral \\   &  with respect to the hypothesis. Only output the label. \\   &  \\   &  Premise: \{premise-demonstration1\}\\   &  Hypothesis: \{hypothesis-demonstration2\}\\   &  Does the premise entail, contradict, is neutral to the hypothesis? \\   &  Answer: \{label1\}\\   &  \textless{}all in-context demonstrations\textgreater\\   &  \\   &  Premise: \{premise-test\}\\   & Hypothesis: \{hypothesis-test\}\\   & Does the premise entail, contradict or is neutral to the hypothesis?\\   & Answer:                                                                                       \\ \bottomrule
\end{tabularx}
\caption{Text of instructions used in ICL. \textsc{ID} and \textsc{NLI}: instructions were adapted from Flan \citep{chung2022-flan} with intent descriptions from the ontology provided with NLU++ \cite{casanueva-etal-2022-nluplusplus}. \textsc{VE}: instructions were adapted from XTREME-UP \cite{ruder-etal-2023-xtremeup} and \citet{ojo2023good}.}\label{tab:icl_instructions}
\end{table}

\section{Finetuning Hyperparameters}
\label{app:hyperparameter_values}

\begin{table}[!h]
\centering
\footnotesize
\begin{tabularx}{0.3\textwidth}{l r}
\toprule
\rowcolor{Gray} Hyperparameter       & Value     \\ \midrule
LaBSE+CL: \textit{dim}           & 512        \\
LaBSE+CL: \textit{non-linearity}           & tanh        \\
Batch size           & 32        \\
Learning rate        & 2e-5      \\
Weight Decay         & 0.1       \\
Evaluation Frequency & 500 steps \\
Max Epochs           & 500       \\
Optimiser            & AdamW     \\ \bottomrule
\end{tabularx}
\caption{Fine-tuning hyperparameters used across supervised training experiments. The rest of the parameters were set to the default values in Huggingface Transformers.}\label{tab:hyperparameter_values}
\end{table}

\section{Full Experimental Results}\label{app:full_experimental_results}

\begin{table*}[!htp]\centering
\footnotesize
\begin{tabularx}{0.95\textwidth}{ll XXXXX XXXXX}\toprule
\rowcolor{Gray} & & \multicolumn{5}{c}{In-domain results} & \multicolumn{5}{c}{Cross-domain results} \\
\rowcolor{Gray} & Samples & \textsc{am} & \textsc{en} & \textsc{mr} & \textsc{es} & \textsc{tr} & \textsc{am} & \textsc{en} & \textsc{mr} & \textsc{es} & \textsc{tr}  \\ \cmidrule(lr){3-7} \cmidrule(lr){8-12}
\multirow{5}{*}{SFT: LaBSE+CL} &30 &0.2998 &0.3253 &0.308 &0.3295 &0.3224 &0.2041 &0.1978 &0.1507 &0.178 &0.1773 \\
&50 &0.3502 &0.3863 &0.3679 &0.4014 &0.3826 &0.2362 &0.2305 &0.1700 &0.2123 &0.1962 \\
&100 &0.4409 &0.5007 &0.4773 &0.4836 &0.4815 &0.2688 &0.2774 &0.2304 &0.2485 &0.2432 \\
&500 &0.6606 &0.7509 &0.7235 &0.7412 &0.7328 &0.4728 &0.5204 &0.4780 &0.4936 &0.5169 \\
&1000 &0.7116 &0.7978 &0.7736 &0.7900 &0.7825 &0.5119 &0.5759 &0.5225 &0.5516 &0.5539 \\ \cmidrule(lr){1-12}
\multirow{5}{*}{SFT: XLM-R} &30 &0.1434 &0.1435 &0.1457 &0.1857 &0.1317 &0.0284 &0.0456 &0.0463 &0.0799 &0.0544 \\
&50 &0.1676 &0.1946 &0.1696 &0.2060 &0.1750 &0.0200 &0.0042 &0.0226 &0.0318 &0.0021 \\
&100 &0.2879 &0.3363 &0.3115 &0.3421 &0.288 &0.1196 &0.1176 &0.0848 &0.1009 &0.1123 \\
&500 &0.5882 &0.742 &0.6592 &0.7075 &0.6898 &0.4107 &0.5076 &0.4441 &0.4694 &0.4806 \\
&1000 &0.6715 &0.8066 &0.7391 &0.7862 &0.7721 &0.4943 &0.5822 &0.5282 &0.5223 &0.5584 \\ \cmidrule(lr){1-12}
\multirow{5}{*}{SIT: Flan-T5-Base} &30 &0.156 &0.6625 &0.1542 &0.4969 &0.2727 &0.0750 &0.5369 &0.0677 &0.4081 &0.1419 \\
&50 &0.1520 &0.7110 &0.1434 &0.5468 &0.3282 &0.0999 &0.5794 &0.0904 &0.4535 &0.1888 \\
&100 &0.1432 &0.7483 &0.1501 &0.6242 &0.3865 &0.1168 &0.6103 &0.0638 &0.4882 &0.2094 \\
&500 &0.1769 &0.8601 &0.1780 &0.7873 &0.6341 &0.1716 &0.7355 &0.1620 &0.6421 &0.4434 \\
&1000 &0.2040 &0.8877 &0.1680 &0.8333 &0.6957 &0.1699 &0.7602 &0.1679 &0.7017 &0.5453 \\ \cmidrule(lr){1-12}
\multirow{5}{*}{SIT: mT0-Base} &30 &0.0560 &0.3735 &0.1558 &0.2646 &0.1505 &0.0125 &0.097 &0.0269 &0.0684 &0.0228 \\
&50 &0.0962 &0.5375 &0.3309 &0.4614 &0.3068 &0.0169 &0.2868 &0.1059 &0.1957 &0.0825 \\
&100 &0.2613 &0.68 &0.5142 &0.6169 &0.516 &0.0936 &0.4795 &0.3009 &0.4209 &0.3151 \\
&500 &0.6488 &0.8222 &0.7466 &0.7978 &0.7579 &0.5394 &0.6711 &0.5985 &0.6441 &0.6163 \\
&1000 &0.6980 &0.8559 &0.7889 &0.8393 &0.8157 &0.5892 &0.7113 &0.6264 &0.6798 &0.6681 \\ \cmidrule(lr){1-12}
ICL: Flan-T5-XL & &0.0328 &0.4927 &0.0302 &0.4526 &0.3136 &0.0554 &0.5375 &0.0581 &0.4176 &0.3012 \\
ICL: mT0-XL & &0.0361 &0.064 &0.0471 &0.0460 &0.0336 &0.0969 &0.0989 &0.0947 &0.1049 &0.1006 \\
ICL: GPT-3.5 & &0.1919 &0.6422 &0.4828 &0.5825 &0.4612 &0.1501 &0.5552 &0.3283 &0.4728 &0.4320 \\
\bottomrule
\end{tabularx}
\caption{Per-language intent detection results for in-domain and cross-domain setups.}
\end{table*}

\begin{table*}[!htp]\centering
\footnotesize
\begin{tabularx}{0.95\textwidth}{ll XXXXX XXXXX}\toprule
\rowcolor{Gray} & & \multicolumn{5}{c}{In-domain results} & \multicolumn{5}{c}{Cross-domain results} \\
\rowcolor{Gray} & Samples & \textsc{am} & \textsc{en} & \textsc{mr} & \textsc{es} & \textsc{tr} & \textsc{am} & \textsc{en} & \textsc{mr} & \textsc{es} & \textsc{tr}  \\ \cmidrule(lr){3-7} \cmidrule(lr){8-12}
\multirow{5}{*}{SFT: XLM-R} &30 &0.1566 &0.2748 &0.1953 &0.2444 &0.2681 &0.0275 &0.0336 &0.0369 &0.03 &0.0232 \\
&50 &0.2199 &0.3234 &0.2603 &0.3221 &0.3098 &0.049 &0.0543 &0.0329 &0.0462 &0.031 \\
&100 &0.4003 &0.4991 &0.3598 &0.4615 &0.4665 &0.0362 &0.0279 &0.0229 &0.0469 &0.0072 \\
&500 &0.6130 &0.7392 &0.6118 &0.6508 &0.6937 &0.036 &0.06 &0.05 &0.103 &0.098 \\
&1000 &0.6468 &0.7801 &0.6614 &0.6855 &0.7539 &0.05 &0.087 &0.083 &0.137 &0.117 \\ \cmidrule(lr){1-12}
\multirow{5}{*}{SIT: Flan-T5-Base} &30 &0.0191 &0.327 &0.0156 &0.2091 &0.1174 &0.0019 &0.2514 &0.0018 &0.07 &0.0511 \\
&50 &0.0362 &0.4486 &0.0083 &0.2627 &0.1537 &0.009 &0.3006 &0.0031 &0.1089 &0.0887 \\
&100 &0.0555 &0.5728 &0.0198 &0.3678 &0.2705 &0.0103 &0.4043 &0.005 &0.1987 &0.1395 \\
&500 &0.0896 &0.7314 &0.042 &0.5073 &0.4615 &0.0313 &0.5956 &0.0141 &0.3689 &0.3272 \\
&1000 &0.1055 &0.8041 &0.0484 &0.5707 &0.5552 &0.0445 &0.6244 &0.014 &0.3975 &0.3577 \\ \cmidrule(lr){1-12}
\multirow{5}{*}{SIT: mT0-Base} &30 &0.1193 &0.3182 &0.1246 &0.2893 &0.1886 &0.0433 &0.1615 &0.0688 &0.1162 &0.1011 \\
&50 &0.1774 &0.3954 &0.1899 &0.347 &0.2488 &0.0511 &0.2153 &0.1118 &0.1679 &0.1371 \\
&100 &0.3313 &0.58 &0.3167 &0.4723 &0.4055 &0.0972 &0.3345 &0.14 &0.212 &0.1927 \\
&500 &0.6093 &0.779 &0.5596 &0.6458 &0.6596 &0.3864 &0.5838 &0.3562 &0.4535 &0.4417 \\
&1000 &0.6521 &0.8193 &0.6036 &0.6814 &0.7065 &0.4304 &0.6528 &0.4337 &0.4572 &0.5097 \\ \cmidrule(lr){1-12}
ICL: Flan-T5-XL & &0.0022 &0.1187 &0.0063 &0.0756 &0.0524 &0 &0.04 &0.02 &0.006 &0.02 \\
ICL: mT0-XL & &0 &0 &0 &0 &0 &0 &0 &0 &0 &0 \\
ICL: GPT-3.5 & &0.1105 &0.3148 &0.1611 &0.2142 &0.1862 & -- & -- & -- & -- & -- \\
\bottomrule
\end{tabularx}
\caption{Per-language value extraction results for in-domain and cross-domain setups.}
\end{table*}

\begin{table}[!htp]\centering
\scriptsize
\begin{tabularx}{0.45\textwidth}{ll XXXX}\toprule
\rowcolor{Gray} & Samples &\textsc{am} & \textsc{mr} & \textsc{es} & \textsc{tr} \\ \cmidrule(lr){1-6}
\multirow{5}{*}{SFT: LaBSE+CL} &30 &0.2123 &0.1742 &0.1869 &0.1574 \\
&50 &0.2595 &0.2076 &0.2392 &0.1959 \\
&100 &0.3586 &0.2977 &0.3287 &0.2693 \\
&500 &0.5821 &0.6027 &0.6465 &0.5704 \\
&1000 &0.6448 &0.7064 &0.7528 &0.6887 \\ \cmidrule(lr){1-6}
\multirow{5}{*}{SFT: XLM-R} &30 &0.0738 &0.1094 &0.1364 &0.1068 \\
&50 &0.0905 &0.1177 &0.1794 &0.1348 \\
&100 &0.1355 &0.2058 &0.2908 &0.2247 \\
&500 &0.2931 &0.4485 &0.6347 &0.5303 \\
&1000 &0.3206 &0.4778 &0.7105 &0.5947 \\ \cmidrule(lr){1-6}
\multirow{5}{*}{SIT: Flan-T5-Base} &30 &0.0385 &0.032 &0.5155 &0.2849 \\
&50 &0.0253 &0.0186 &0.5567 &0.2959 \\
&100 &0.0218 &0.023 &0.5858 &0.3065 \\
&500 &0.0449 &0.0469 &0.672 &0.3575 \\
&1000 &0.0947 &0.0993 &0.692 &0.3585 \\ \cmidrule(lr){1-6}
\multirow{5}{*}{SIT: mT0-Base} &30 &0.0923 &0.1901 &0.3008 &0.1889 \\
&50 &0.194 &0.3336 &0.4732 &0.3298 \\
&100 &0.3078 &0.4534 &0.5898 &0.4716 \\
&500 &0.4305 &0.5893 &0.7329 &0.6316 \\
&1000 &0.4602 &0.6246 &0.765 &0.6638 \\ \cmidrule(lr){1-6}
ICL: Flan-T5-XL & &0.0633 &0.0667 &0.447 &0.2804 \\
ICL: mT0-XL & &0.0719 &0.0799 &0.0723 &0.0645 \\
ICL: GPT-3.5 & &0.1487 &0.3867 &0.5764 &0.475 \\
\bottomrule
\end{tabularx}
\caption{Per-language intent detection results for the cross-lingual setup for \textsc{en} $\rightarrow$ \textsc{tgt} transfer.}
\end{table}

\begin{table}[!htp]\centering
\scriptsize
\begin{tabularx}{0.45\textwidth}{ll XXXX}\toprule
\rowcolor{Gray} & Samples &\textsc{am} & \textsc{mr} & \textsc{es} & \textsc{tr} \\ \cmidrule(lr){1-6}
\multirow{5}{*}{SFT: XLM-R} &30 &0.0867 &0.11 &0.2004 &0.1531 \\
&50 &0.1206 &0.1623 &0.2511 &0.213 \\
&100 &0.1748 &0.2151 &0.3278 &0.2847 \\
&500 &0.3174 &0.405 &0.5335 &0.4647 \\
&1000 &0.3569 &0.4256 &0.5769 &0.5177 \\ \cmidrule(lr){1-6}
\multirow{5}{*}{SIT: Flan-T5-Base} &30 &0.0567 &0.0299 &0.2122 &0.1032 \\
&50 &0.0481 &0.0265 &0.238 &0.1236 \\
&100 &0.0531 &0.0316 &0.2885 &0.1411 \\
&500 &0.0659 &0.0422 &0.3549 &0.1706 \\
&1000 &0.0806 &0.0435 &0.3837 &0.1788 \\ \cmidrule(lr){1-6}
\multirow{5}{*}{SIT: mT0-Base} &30 &0.1045 &0.0872 &0.2471 &0.1728 \\
&50 &0.1123 &0.0994 &0.2937 &0.1895 \\
&100 &0.1447 &0.1556 &0.3939 &0.2382 \\
&500 &0.1855 &0.2141 &0.5445 &0.3293 \\
&1000 &0.2049 &0.2327 &0.5585 &0.3394 \\ \cmidrule(lr){1-6}
ICL: Flan-T5-XL & &0 &0.001 &0.03 &0 \\
ICL: mT0 & &0 &0 &0 &0 \\
ICL: GPT-3.5 &   & -- & -- & -- & -- \\
\bottomrule
\end{tabularx}
\caption{Per-language value extraction results for the cross-lingual setup for \textsc{en} $\rightarrow$ \textsc{tgt} transfer.}
\end{table}

\begin{table}[!htp]\centering
\scriptsize
\begin{tabularx}{0.45\textwidth}{ll XXXX}\toprule
\rowcolor{Gray} & Samples &\textsc{ru} & \textsc{es} & \textsc{en} & \textsc{tr} \\ \cmidrule(lr){1-6}
\multirow{5}{*}{SFT: XLM-R} &30 &0.3426 &0.333 &0.3422 &0.3373 \\
&50 &0.3554 &0.3357 &0.3538 &0.3285 \\
&100 &0.362 &0.3321 &0.3578 &0.3345 \\
&500 &0.4458 &0.4482 &0.5378 &0.4357 \\
&1000 &0.545 &0.4795 &0.5936 &0.4663 \\ \cmidrule(lr){1-6}
\multirow{5}{*}{SIT: Flan-T5-Base} &30 &0.4399 &0.5094 &0.6255 &0.385 \\
&50 &0.4355 &0.5106 &0.6844 &0.4212 \\
&100 &0.4577 &0.5451 &0.7623 &0.3914 \\
&500 &0.5409 &0.6537 &0.807 &0.5339 \\
&1000 &0.5549 &0.6593 &0.8238 &0.5423 \\ \cmidrule(lr){1-6}
\multirow{5}{*}{SIT: mT0-Base} &30 &0.3545 &0.388 &0.3878 &0.3641 \\
&50 &0.4034 &0.3756 &0.4762 &0.3567 \\
&100 &0.4551 &0.4321 &0.4848 &0.402 \\
&500 &0.5427 &0.5563 &0.615 &0.5238 \\
&1000 &0.5581 &0.6132 &0.6403 &0.5425 \\ \cmidrule(lr){1-6}
ICL: Flan-T5-XL & &0.6307 &0.7808 &0.8994 &0.5758 \\
ICL: mT0 & &0.33 &0.33 &0.33 &0.33 \\
ICL: GPT-3.5 & &0.5333 &0.5683 &0.6224 &0.518 \\
\bottomrule
\end{tabularx}
\caption{Per-language natural language inference results on XNLI.}
\end{table}

\section{Further Value Extraction Results}\label{app:other_ve_results}

\begin{figure*}[!h]
\centering
\begin{subfigure}[b]{0.95\textwidth}
   \includegraphics[width=1\linewidth]{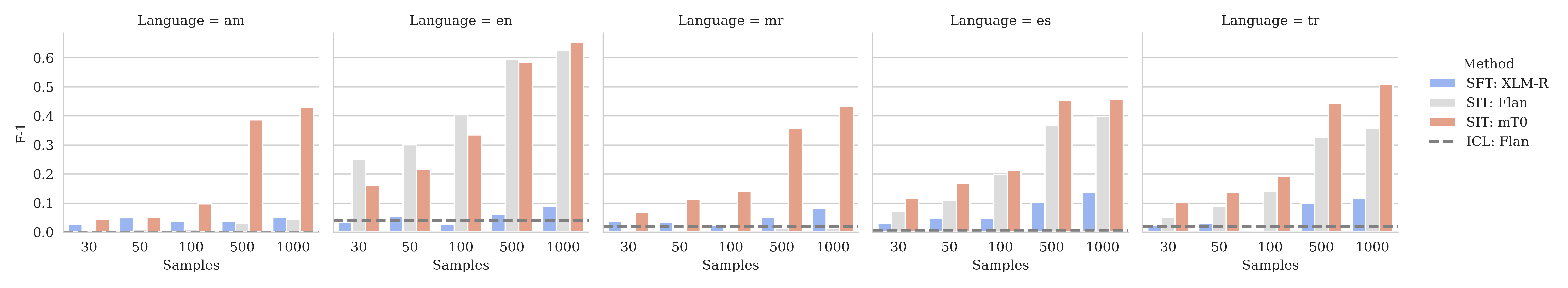}
   \caption{\textsc{VE}: Cross-Domain In-Language}
   \label{fig:VE_crossdomain_inlanguage} 
\end{subfigure}

\begin{subfigure}[b]{0.90\textwidth}
   \includegraphics[width=1\linewidth]{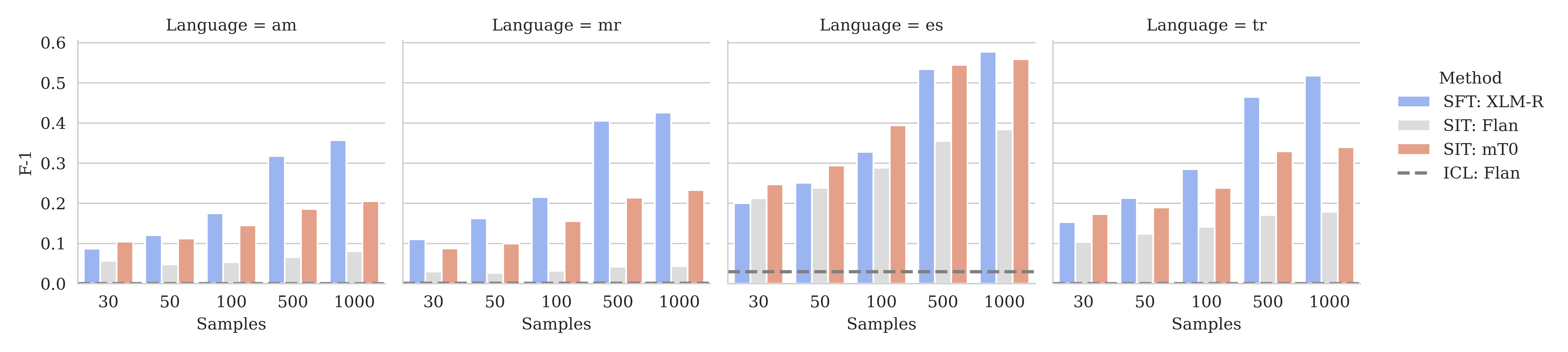}
   \caption{\textsc{VE}: Cross-Lingual In-Domain}
   \label{fig:VE_indomain_crosslingual}
\end{subfigure}

\caption{Value extraction results for Amharic (\textsc{am}), English (\textsc{en}), Marathi (\textsc{mr}), Spanish (es) and Turkish (tr) for two setups: a) cross-domain in-language; and b) cross-lingual in-domain performance. We exclude ICL-mT0 XL from the plot, as it had 0.0 performance on VE task in these setups. Qualitative analysis of the outputs of ICL-mT0 showed that the outputs neither adhered to the slot-value pair formatting nor included the right values.} \label{app_fig:ve_few_shot_results}
\end{figure*}

\clearpage
\section{QLoRA Finetuning Hyperparameters}\label{app:qlora_hyperparameters}

\begin{table}[!h]
\centering
\footnotesize
\begin{tabularx}{0.3\textwidth}{l r}
\toprule
\rowcolor{Gray} Hyperparameter       & Value     \\ \midrule
\multicolumn{2}{c}{\textit{Quantisation}} \\ \midrule
Precision           &   4-bit     \\
dtype        & float16 + nf4      \\ \midrule
\multicolumn{2}{c}{\textit{LoRA}} \\ \midrule
Rank & 64 \\
$\alpha$           & 16       \\
Dropout          & 0.1       \\
LoRA Modules            & All Layers     \\ \midrule
\multicolumn{2}{c}{\textit{Finetuning}} \\ \midrule
Scheduler            & Constant     \\ 
Learning Rate            & 0.0002     \\ 
Warm-Up Ratio            & 0.03     \\
Batch Size            & 16     \\
Weight Decay            & 0.0     \\
Learning Steps            & 10000     \\
\bottomrule
\end{tabularx}
\caption{Hyperparameters for QLoRA tuning. Except for the hyperparameters provided, the rest were set to the default values in Huggingface Transformers.}\label{tab:hyperparameter_values_qlora}
\end{table}

\section{In-Context Learning Results for XNLI}\label{app:xnli_results}

\setlength{\tabcolsep}{4.5pt}
\begin{table}[!h]
\def\arraystretch{0.81}
{\fontsize{6.9}{7.1}\selectfont
%\footnotesize
%\resizebox{\linewidth}{!}{%
\centering
\begin{tabularx}{0.99\linewidth}{ll XXXX}
\toprule
\multirow{1}{*}{\em Model} &  & \textsc{es} & \textsc{ru} & \textsc{tr} & \textsc{avg}                               \\ \midrule
\multirow{2}{*}{\textit{LLaMA-2-7B}}  & Raw                     &     32.3    & 33.1 & 26.2 &     30.5        \\
                     &  +QLoRA                    &   33.1 & 33.4  & 25.8 &          30.7         \\
 \bottomrule
\end{tabularx}
}
%}
\vspace{-1mm}
\caption{ICL results for XNLI before and after QLoRA language adaptation. } \label{app_tab:xnli_icl_qlora}
\vspace{-2mm}
\end{table}

\newpage
\section{Human Annotation Instructions for Naturalness and Usefulness}\label{app:annotation_instructions}

\begin{table}[!h]
\footnotesize
\begin{tabularx}{0.45\textwidth}{lX}
\toprule
\textbf{Score} & \textbf{Instruction Text}   \\  \midrule
\multicolumn{2}{c}{\textit{Naturalness}} \\ \midrule
\multicolumn{2}{l}{Please read the text below and evaluate its \textit{naturalness}} \\ 
\multicolumn{2}{l}{using 3-point scale below:} \\
1 & The text is not natural, you would never say anything similar as a native speaker.\\

2 & The text is generally natural but contains some elements which feel odd.\\

3 & The text is completely natural, it contains nothing you find odd.\\ \midrule
 \multicolumn{2}{c}{\textit{Usefulness}} \\ \midrule
 \multicolumn{2}{l}{Please read the instruction and the text and evaluate} \\ 
\multicolumn{2}{l}{its \textit{usefulness} using 3-point scale below:} \\
1 &   The text does not provide any useful information to complete the task in the instruction.\\

2 & The text contains some information which could be useful for the instructions.\\

3 & The text contains full information and completes the task in the instructions. \\  
  \bottomrule
\end{tabularx}
\caption{Text of instructions used in human evaluation.}
\end{table}

\end{document}